\documentclass[lettersize,journal]{IEEEtran}
\usepackage{amsmath,amsfonts,amsthm,amssymb}
\usepackage{algorithm,algorithmicx}
\usepackage{caption}
\usepackage{algcompatible}
\usepackage{array}
\usepackage[caption=false,font=normalsize,labelfont=sf,textfont=sf]{subfig}
\usepackage{textcomp}
\usepackage{stfloats}
\usepackage{url}
\usepackage{verbatim}
\usepackage{graphicx}
\usepackage{cite}
\usepackage[colorlinks,
            linkcolor=black,
            anchorcolor=black,
            citecolor=black
            ]{hyperref}
\usepackage{booktabs}
\newcommand{\ra}[1]{\renewcommand{\arraystretch}{#1}}
\usepackage[noend]{algpseudocode}

\newtheorem{technique}{Technique}
\newtheorem{proposition}{Proposition}

\begin{document}

\title{Nested Annealed Training Scheme for Generative Adversarial Networks}

\author{Chang Wan, Ming-Hsuan Yang,~\IEEEmembership{Fellow,~IEEE,} Minglu Li,~\IEEEmembership{Fellow,~IEEE,}, Yunliang Jiang,  Zhonglong Zheng,~\IEEEmembership{Member,~IEEE,}
\thanks{Manuscript received on February 17, 2024, and first revised on August 24, 2024. This work was funded by the National Natural Science Foundation of China U22A20102, 62272419, Natural Science Foundation of Zhejiang Province ZJNSFLZ22F020010.}
\thanks{Chang Wan, Minglu Li, Yunliang Jiang, Zhonglong Zheng are with the
School of Computer Science and Technology, Zhejiang Normal University,
Jinhua 321004, China (e-mail: wanchang\_phd2020@zjnu.edu.cn, mlli@zjnu.edu.cn, jyl2022@zjnu.cn, zhonglong@zjnu.edu.cn).}

\thanks{Ming-Hsuan Yang is with the department of Computer Science and Engineering at University of California, Merced, 95344, California, USA. (email: mhyang@ucmerced.edu)}}

\markboth{Journal of \LaTeX\ Class Files,~Vol.~14, No.~8, August~2021}%
{Shell \MakeLowercase{\textit{et al.}}: A Sample Article Using IEEEtran.cls for IEEE Journals}


\maketitle
\begin{abstract}
Recently, researchers have proposed many deep generative models, including generative adversarial networks (GANs) and denoising diffusion models.
Although significant breakthroughs have been made and empirical success has been achieved with the GAN, its mathematical underpinnings remain relatively unknown. 
This paper focuses on a rigorous mathematical theoretical framework: the composite-functional-gradient GAN (CFG)~\cite{johnson2018composite}.
Specifically, we reveal the theoretical connection between the CFG model and score-based models.
We find that the CFG discriminator's training objective is equivalent to finding an optimal $D(\mathbf{x})$. The optimal $D(\mathbf{x})$'s gradient differentiates the integral of the differences between the score functions of real and synthesized samples. 
Conversely, training the CFG generator involves finding an optimal $G(\mathbf{x})$ that minimizes this difference. 
In this paper, we aim to derive an annealed weight preceding the CFG discriminator's weight.
This new explicit theoretical explanation model is called the annealed CFG method.
To overcome the annealed CFG method's limitation, as the method is not readily applicable to the state-of-the-art (SOTA) GAN model, we propose a nested annealed training scheme (NATS). 
%
This scheme keeps the annealed weight from the CFG method and can be seamlessly adapted to various GAN models, no matter their structural, loss, or regularization differences.
We conduct thorough experimental evaluations on various benchmark datasets for image generation. The results show that our annealed CFG and NATS methods significantly improve the synthesized samples' quality and diversity. 
This improvement is clear when comparing the CFG method and the SOTA GAN models.

\end{abstract}


\maketitle
\section{Introduction}
Recently, many deep generative models have surfaced across diverse application domains, including image and video synthesis~\cite{ho2022imagen, ramesh2021zero, ramesh2022hierarchical, pan2023drag}, three-dimensional (3D) object creation~\cite{stan2023ldm3d, zheng2023locally}, and audio generation~\cite{ruan2023mm}. 
Specifically, generative models---including generative adversarial networks (GANs)\cite{goodfellow2014generative} and the diffusion model (DM)---have been widely used for image synthesis and related tasks. 
GAN works by sampling a low-dimensional vector set. These vectors are then input into a neural network to generate synthetic images through a single forward propagation. 
This framework is known for its capability to synthesize highly realistic images and its fast sampling process. However, some details require more in-depth analysis~\cite{10378665,10012343,9953153}. 
In addition to its broad application prospects, GAN is known for its mode collapse and non-convergent training issues. 
Plenty of GAN literature exists from the theoretical perspective to analyze these training problems. 
Previous research on GAN's theoretical explanation can be categorized into the following four broad areas: game theory, optimization or dynamic theory, metrics between distributions, and generalization error analysis. 
The game theory extends GAN's two-player zero-sum game to include more game modes and multiple players from a theoretical perspective.
Nevertheless, these game theory methods' empirical results and generalizations are unsatisfactory.
The optimization and dynamic theory focuses on the discriminator's and generator's first-order or second-order gradient during training, attempting to reach a global optimal point for GAN. 
That said, these theories' assumptions and conditions are often overly strict and fail to accurately reflect real-world data. As such, the numerical settings of these methods do not align well with the theoretical analysis, producing sub-optimal results.
The metric theory concentrates on the new metric between real and synthesized data distribution, which is practical when working with different data distributions from a theoretical perspective. 
However, these metrics' theoretical formula is difficult to compute in the empirical process, so it is estimated using hyper-parameters, producing sub-optimal results.
Analyzing generalization errors allows us to identify the key factors and their mathematical relationships for GAN generalization in machine-learning theory. 
However, these analyses do not offer practical GAN training guidance.
Theoretical research on GAN has generally involved controlling the discriminator's gradients to help guide the generator. 
That said, there is not much clarity regarding the theoretical formula for the discriminator's gradients and how the discriminator influences the generator's training.
The above content is explained in detail in Appendix Section E.

This paper analyzes the gradient of the discriminator's theoretical formula using the score function. 
Our aim with this work is to better understand the GAN model.
%
We concentrate on the composite-functional-gradient GAN (CFG), which has a rigorous mathematical foundation.  
This model distinguishes the training processes of the discriminator and generator, enabling us to gain a deeper understanding of GAN's underlying principles.
%
The CFG discriminator's gradient $\nabla_{\mathbf{x}} D(\mathbf{x})$ computes the difference between the score functions of real and synthesized samples and then integrates it.
Conversely, the optimal CFG generator $G(\mathbf{x})$ minimizes this difference.
This approach helps reveal a connection between the CFG and score-based models, establishing a theoretical framework for understanding GAN.
%
Furthermore, we derive an annealed weight preceding the CFG discriminator's gradient. 
We design this adjustment to align with score-based models' theoretical requirements. 
We call the proposed method the annealed CFG method.


%
%
Note that applying the annealed CFG method to train state-of-the-art (SOTA) GAN models is difficult owing to intrinsic differences in the training process. 
This paper proposes a nested annealed training scheme (NATS) that introduces the CFG model's annealed weight and can be adapted to a wider GAN model spectrum. 
From the dynamic-theory perspective, we demonstrate that the NATS' gradient-vector field is equal to that of annealed CFG. 
Empirically, our NATS methodology shows adaptability across various GAN models, no matter their structural, loss, or regularization differences. 
Moreover, extensive experimental results show that SOTA GAN models achieve significant performance gains when trained using our NATS approach. 

This work's main contributions are as follows: 
\begin{itemize}

\item We give the theoretical formula for $\nabla_{\mathbf{x}} D(\mathbf{x})$ in CFG, a GAN model type, and identify its theoretical connection to score-based models.
The CFG discriminator's gradient $\nabla_{\mathbf{x}} D(\mathbf{x})$ computes the difference between the score functions of real and synthesized samples and integrates it.
Conversely, the optimal CFG generator $G(\mathbf{x})$ minimizes this difference;



\item Referring to score function theory, we establish an annealed weighting scheme from the CFG method, which can be applied to GAN models. 
We show that the NATS training-scheme generator's gradient-vector field is equal to that of the annealed CFG method;


\item Empirically, we show the CFG method' and NATS' effectiveness.
Our annealed CFG and NATS also achieve favorable performance in quality and diversity compared with the CFG method and typical GAN training schemes. 
Notably, when trained using NATS, the SOTA GAN models make significant performance gains over prior SOTA models. 
\end{itemize}

\begin{algorithm}[htb]
\caption{Typical training schemes of Generative Adversarial Networks. 
\label{gan}} 
\begin{algorithmic}[1]
\For{$i=1,2, \ldots, N$}
\For{$k=1,2, \ldots, K$}
\State
Sample minibatch samples $\left\{\mathbf{z}^{(1)}, \ldots, \mathbf{z}^{(m)}\right\}$ from  \Statex $\qquad\ \ $ noise prior $p_{\mathbf{z}}$
\State
Sample minibatch samples $\left\{\mathbf{x}^{(1)}, \ldots, \mathbf{x}^{(m)}\right\}$ from  
\Statex $\qquad\ \ $ data distribution  $p_{\text {data }}(\mathbf{x})$  
\State
Update the discriminator by \textbf{ascending} its 
\Statex $\qquad\ \ $ stochastic gradient:
\Statex \begin{footnotesize}
$\qquad\qquad$ 
$
\nabla_{\theta_d} \frac{1}{m} \sum_{i=1}^m\left[\log D\left(\mathbf{x}^{(i)}\right)+\log \left(1-D\left(G\left(\mathbf{z}^{(i)}\right)\right)\right)\right]
$
\end{footnotesize}
\EndFor
\State {\bf end for}
\State
Sample minibatch $\left\{\mathbf{z}^{(1)}, \ldots, \mathbf{z}^{(m)}\right\}$ from $p_{\mathbf{z}}$ 
\State
Update the generator by \textbf{descending} its stochastic 
\Statex $\quad\;$ gradient:\begin{footnotesize} 
$
\nabla_{\theta_g} \frac{1}{m} \sum_{i=1}^m \log \left(1-D\left(G\left(\mathbf{z}^{(i)}\right)\right)\right) .
$
\end{footnotesize}
\EndFor
\State {\bf end for}
\end{algorithmic}
\end{algorithm}

\section{Preliminary}

\vspace{1mm}
\noindent \textbf{GANs.}
GAN is a framework for one-step generative models made up of two interconnected deep neural networks: the discriminator and the generator. 
The generator represented as $G( \mathbf{z};\theta_g)$, works as a mapping mechanism that transforms a prior distribution $p_{\mathbf{z}}$ into an approximation of the real samples' distribution $p_g$ through a deep neural network. 
Conversely, the discriminator, $D( \mathbf{x};\theta_d)$, maps data from $p_g$ or the real samples' distribution $p_{data}$ 
to a scalar value. 
The discriminator tries to enhance its ability to accurately label real examples and samples synthesized by $G$. 
In contrast, the generator attempts to reduce the discriminator's accuracy in labeling, thus minimizing the probability of correct classification by $G$.
We explain the philosophy behind this idea in Algorithm~\ref{gan}.

\vspace{1mm}
\noindent \textbf{CFG.}
CFG employs a discriminator that functions as a logistic regression to distinguish between real and synthetic samples~\cite{johnson2019framework}.
Moreover, it uses the functional compositions gradient to learn the generator in the following form to obtain $G(\mathbf{z})=G_{M}(\mathbf{z})$:
\begin{small}
\begin{gather}
   \begin{aligned}
  G_{m}( \mathbf{z})=G_{m-1}( \mathbf{z})+\eta_{m}g_m\left(G_{m-1}(\mathbf{z})\right), m=1, \ldots,M \label{eq:generator}
\end{aligned},
\end{gather}
\end{small}
where $M$ represents the number of steps in the generator, which are used to approximate real data samples' distribution, and each $g_m$ is a function estimated from data.

The aim is to learn $g_m: \mathbb{R}^k \rightarrow \mathbb{R}^k$ from data such that the probability density $p_m$ of $G_m(\mathbf{z})$, which continuously changes by ~(\ref{eq:generator}) and becomes close to the density $p_*$ of real data as $m$ continuously increases. 
To measure the closeness, we employ
$L$, denoting a distance measure such as the KL divergence between two distributions. 
From the following equation, we obtain the choice of $g_m(\cdot)$ that guarantees that transformation~(\ref{eq:generator}) can always reduce/improve $L(\cdot)$.
Let $\ell_2^{\prime}\left(p_*, p_m\right)=$ $\partial \ell\left(p_*, p_m\right) / \partial p_m$. 
Thus, we have the following:
\begin{small}
\begin{gather}\label{eq:theorem 2.1}
    \begin{aligned}
       \frac{d L\left(p_m\right)}{d m}=\int p_m( \mathbf{x}) \nabla_{\mathbf{x}} \ell_2^{\prime}\left(p_*( \mathbf{ \mathbf{x}}), p_m( \mathbf{x})\right) \cdot g_m( \mathbf{x}) d x \leq 0.
    \end{aligned}
\end{gather}
\end{small}
The above is the most crucial theoretical theorem in CFG, indicating that the distance between synthetic and real samples will decrease until $0$.
With this definition of $\frac{d L\left(p_m\right)}{d m}$, we want to keep $\frac{d L\left(p_m\right)}{d m}$ as negative so that the distance $L$ decreases.
To achieve this goal, we set $g_m(x)$ as
\begin{gather}
\begin{aligned}
  g_m(\boldsymbol{x})=-s_m(\boldsymbol{x}) \phi_0\left(\nabla_x \ell_2^{\prime}\left(p_*(\boldsymbol{x}), p_m(\boldsymbol{x})\right)\right),
\end{aligned}\label{eq:gm}
\end{gather}
where $s_m(x)>0$ is a random scaling factor. 
$\phi_0(u)$ is a vector function such that $\phi(u)=u \cdot \phi_0(u) \geq 0$ and $\phi(u)=0$ if and only if $u=0$. 
Two examples are $(\phi_0(u)=u, \phi(u)=\|u\|_2^2)$ and $(\phi_0(u)=\operatorname{sign}(u), \phi(u)=\|u\|_1)$.
Empirically, we define the function in ~(\ref{eq:gm}) as 
$g_m( \mathbf{x})=\delta_m(\mathbf{x}) \nabla_{\mathbf{x}} D( \mathbf{x}) $, with  $\delta_m(\mathbf{x})$ computed as 
$
\delta_m(\mathbf{x})=s_m( \mathbf{x})\tilde{k}_m( \mathbf{x})f_{kl}''(\tilde{k}_m( \mathbf{x})) \label{eq:gx1}
$, 
where we have the arbitrary scaling factor $s_m(\mathbf{x})$, the KL-divergence function $f_{kl}=-\ln  \mathbf{x}$, and $\tilde{k}_m(\mathbf{x})=\exp (-D(\mathbf{x})) \approx \frac{p_{g_m}(\mathbf{x})}{p_*(\mathbf{x})}=k_m(\mathbf{x})$.
\begin{figure}[tp]
\centering 
\includegraphics[width=0.45\textwidth]{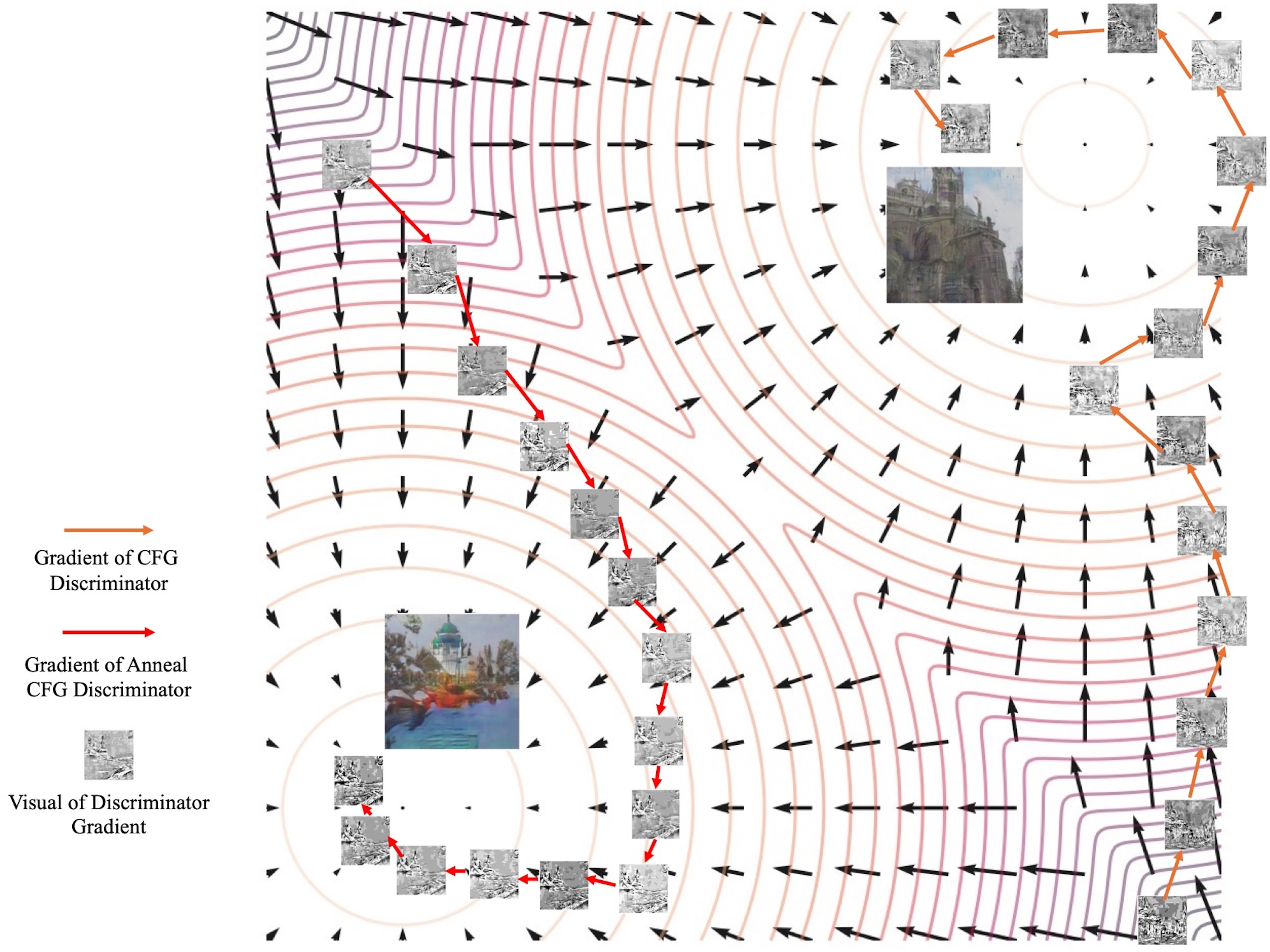}
\caption{
Intuitive understanding of our annealed weight mechanisms. 
\label{Fig.Intuitive}}
\vspace{-0.2in}
\end{figure}

\vspace{1mm}
\noindent \textbf{Score-Based Model.}
The score-based model is a multi-step generative denoising diffusion model framework.
A neural network $\mathbf{s}_{\theta}(\mathbf{x})$ is established to estimate the $\nabla_{\mathbf{x}}\log p_{\text {data }}(\mathbf{x})$ of the real data samples in each diffusion step. 
For a given $\sigma$, the denoising score matching objective is as follows:
$$
\ell(\boldsymbol{\theta} ; \sigma) \triangleq \frac{1}{2} \mathbb{E}_{p_{\text {data }}(\mathbf{x})} \mathbb{E}_{\tilde{\mathbf{x}} \sim \mathcal{N}\left(\mathbf{x}, \sigma^2 I\right)}\left[\left\|\mathbf{s}_{\boldsymbol{\theta}}(\tilde{\mathbf{x}}, \sigma)+\frac{\tilde{\mathbf{x}}-\mathbf{x}}{\sigma^2}\right\|_2^2\right],
$$ where $\boldsymbol{\theta}$ represents the parameter of the neural network $\mathbf{s}_{\theta}(\mathbf{x})$, $\mathbf{x}$ denotes the real data samples, and $\tilde{\mathbf{x}}$ signifies real data samples with added noise of variance $\sigma^2$.
We can combine the above equations for all $\sigma \in\left\{\sigma_i\right\}_{i=1}^L$ to obtain one unified objective:
$$
\mathcal{L}\left(\boldsymbol{\theta} ;\left\{\sigma_i\right\}_{i=1}^L\right) \triangleq \frac{1}{L} \sum_{i=1}^L \lambda\left(\sigma_i\right) \ell\left(\boldsymbol{\theta} ; \sigma_i\right),
$$
where $\lambda\left(\sigma_i\right)>0$ is a coefficient function depending on $\sigma_i$.
Then, the score-based model uses the Langevin dynamics sampling to extract samples from the  $\mathbf{s}_{\theta}(\mathbf{x})$.  
It initializes the chain from an arbitrary prior distribution $\mathbf{x}_0 \sim 
\pi(\mathbf{x})$ and then iterates the following:
\begin{gather}\label{sampling}
\mathbf{x}_{t+1} \leftarrow \mathbf{x}_t+\epsilon \mathbf{s}_{\theta}(\mathbf{x})+\sqrt{2 \epsilon} \mathbf{z}_t, \quad t=0,1, \cdots, K,
\end{gather}
where $\mathbf{z}_t \sim \mathcal{N}(0, I)$. 
When $\epsilon \rightarrow 0$ and $K \rightarrow \infty, \mathbf{x}_K$, obtained from the above procedure, converges to a real sample from $p_{\text {data }}(\mathbf{x})$ under certain regularity conditions.
However, in practice, the error is trivial when $\epsilon$ is sufficiently small and $K$ is sufficiently large.
\begin{figure}[h]
\centering 
\includegraphics[width=0.48\textwidth]{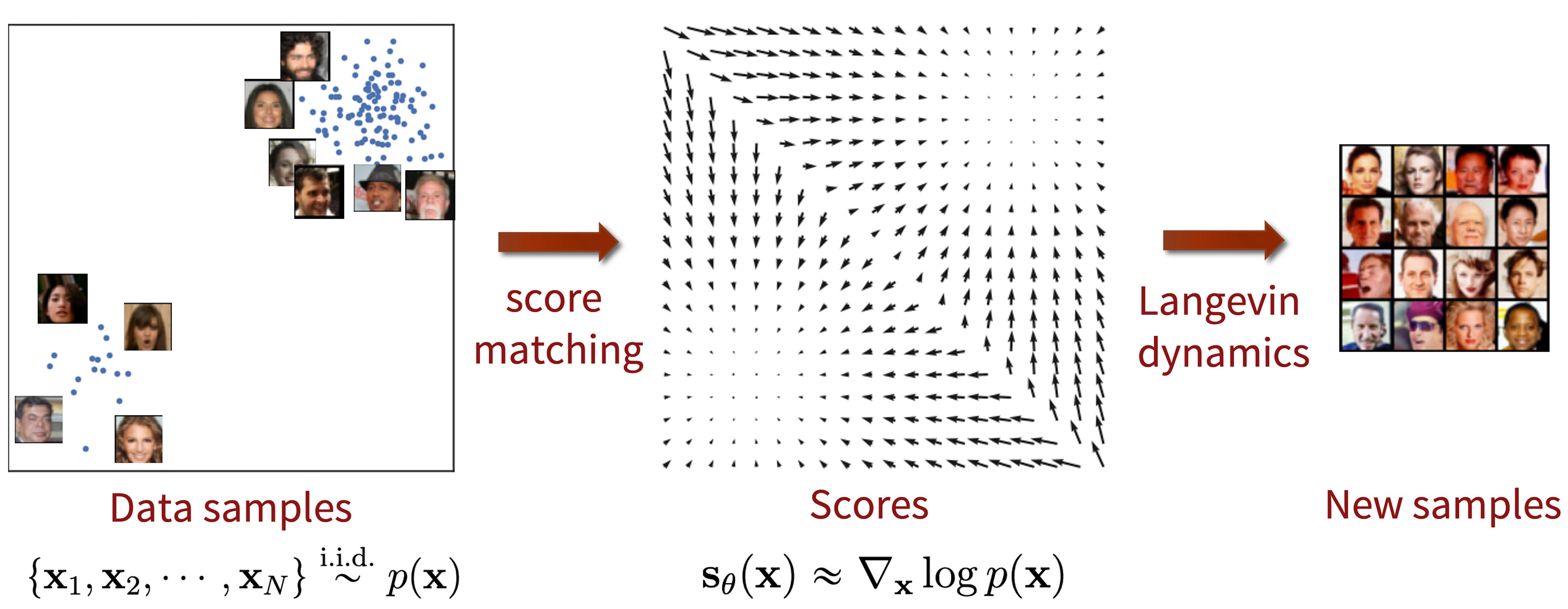}
\caption{
The figure illustrates the concept derived from a score-based model~\cite{song2019generative}, where the score represents the gradient of the logarithmic density function of real samples. This indicates the direction through which noise samples evolve towards real samples.
\label{Fig.motivation}}
\end{figure}

\noindent \textbf{Annealed Langevin Dynamics Sampling.}
Annealed Langevin dynamics sampling, or annealed sampling for short, is an enhancement of Langevin dynamics sampling. 
Similarly to the ~(\ref{anneal-sampling}), annealed sampling is a method that combines information from all
noise scales and builds an annealed weight for $\mathbf{s}_{\theta}(\mathbf{x})$:
\begin{gather}\label{anneal-sampling}
\mathbf{x}_{t+1} \leftarrow \mathbf{x}_t+\alpha_i\mathbf{s}_{\theta}(\mathbf{x})+\sqrt{2 \epsilon} \mathbf{z}_t, \quad t=0,1, \cdots, K,
\end{gather}
where $\alpha_i= \epsilon \cdot \sigma_i^2 / \sigma_L^2$, and $\epsilon$ is a step size, as in ~(\ref{anneal-sampling}).
The noise scale $\left\{\sigma_i\right\}_{i=1}^L$ satisfies $\sigma_1>\sigma_2>\cdots>\sigma_L$.
Choose $\left\{\sigma_i\right\}_{i=1}^L$ as a geometric progression with the common ratio $\gamma$ such that $\Phi(\sqrt{2 D}(\gamma-1)+3 \gamma)-\Phi(\sqrt{2 D}(\gamma-1)-3 \gamma) \approx 0.5$, where $\Phi(.)$ is a CDF of standard Gaussian. 
With this chosen, the value $\alpha_i$ represents a descending order such that $\alpha_1>\alpha_2>\cdots>\alpha_L$. For this reason, the method is called the annealed Langevin dynamics sampling.

\section{Method}
This section provides an overview of the proposed method before describing the details and relationships to prior approaches. 
\subsection{Overview}\label{overview}
\noindent\textbf{Intuition.}
Fig.~\ref{Fig.Intuitive} shows the annealing approach's main idea. 
The lower-left image shows superior visual samples, while the upper-right image shows inferior samples, set against the backdrop of a real data distribution's score function.
At first, the distribution derives from a noise distribution (the blank diagonal area).
Focusing on the right orange lines from the bottom right starting point, we can see that the uniform length of these lines symbolizes the weight of each discriminator gradient. 
The gray block represents the discriminator gradient's visual representation. 
However, this constant weight at each step inadvertently leads to sub-optimal samples, bypassing the ideal ones.

Conversely, examining the left red lines starting from the top left, we see that the length of these lines gradually decreases, representing the diminishing weight of each discriminator gradient. 
The gray block shows the discriminator gradient's visual representation. As training proceeds, the discriminator gradient methodically follows the score function, finally converging on the real samples.

This situation can be more clearly viewed through the lens of optimization. 
At the optimization's outset, when the starting point is far from the optimum, the weight assigned to each step can be made larger to speed up the search.
However, as the search point nears the optimum, reducing the weight of each step becomes prudent to ensure convergence. 
Veering from this annealing principle can cause the search to deviate from the optimum, producing inferior results. 
This behavior parallels the scenarios in both the CFG and annealed CFG methods.
Our demonstration that the discriminator's gradient is correlated with the score function verifies these intuitive concepts.

\vspace{1mm}
\noindent\textbf{Notation.} We distinguish between the notation integral and our theoretical proof below.
$D(\mathbf{x};\theta_d):\mathcal{X} \rightarrow \mathcal{C}$, abbreviated as $D(\mathbf{x})$, represents the discriminator characterized by the parameter $\theta_d$.
$G(\mathbf{z};\theta_g):\mathcal{Z} \rightarrow \mathcal{X}$, abbreviated as $G(\mathbf{z})$, is the generator with the parameter $\theta_g$.
The vector $\mathbf{w}$ represents the annealed weight vector pertinent to the gradient of the discriminator.
The notations $\mathbf{w}(\mathbf{x})$ 
represent the annealed weight vector.
$\mathbf{w}_{m}(\mathbf{x})$ and $\mathbf{w}[m]$ both indicate the $m$-th element of the $\mathbf{w}(\mathbf{x})$ vector.
The pairwise distance between samples in datasets is denoted by $D$. 
$D_{1}$ is the median pairwise distance between training data, and $D_{2}$ refers to the median pairwise distance between training and synthesized samples.

\vspace{1mm}
\noindent\textbf{Definition.}
This paper introduces a novel training framework: the NATS. 
We refer to the nested training scheme without the annealed weight as \textbf{NTS} for a comparison. 
We then contrast these schemes with the traditional training approach in GANs, referred to as the common training scheme (\textbf{CTS}), detailed in Algorithm~\ref{gan}. 
The term $N_d$ denotes the nested number specified in Algorithm~\ref{natsgan}, which is integral to our proposed methodologies.

\vspace{1mm}
\noindent\textbf{Proof.}
The proof supporting this section's proposition is detailed in Appendix Section B.

\subsection{CFG And Score Function Interrelation}\label{cfg and score}
Previous work on GAN models has focused on controlling the discriminator's gradients and enabling the generator to produce high-quality samples. 
However, there is a lack of clarity in the discriminator's gradients' theoretical formula and how it influences the generator's training.
This section discusses how to create the discriminator gradients using the score function and how it guides the generator's training.

\vspace{1mm}
\noindent\textbf{Interrelation of CFG Discriminator and Score Function.
}
Upon scrutinizing the discriminator in the CFG method's loss function, we find that
\begin{small}
\begin{gather}\label{cfg:discriminator loss}
\begin{aligned}
    \mathop{\min}_{D} \left[\mathbf{E}_{x \sim p_*} \ln \left(1+e^{-D(\mathbf{x})}\right)+\mathbf{E}_{x \sim p_g} \ln \left(1+e^{D(\mathbf{x})}\right)\right]
\end{aligned}
\end{gather}
\end{small}
and its analytical solution formula
\begin{gather}\label{cfg:solution}
\begin{aligned}
D(\mathbf{x}):=\log \frac{p_*(\mathbf{x})}{p_g(\mathbf{x})}\end{aligned},
\end{gather}
and we propose the proposition below based on this analytic solution formula.

\begin{proposition}\label{p:1}
Letting $p_*(\mathbf{x})$ and $p_g(\mathbf{x})$ denoting the distribution of real samples and synthesis samples, respectively, we obtain
$$
\nabla_{\mathbf{x}} D(\mathbf{x}) = \nabla_{\mathbf{x}} \log p_*(\mathbf{x}) -  \nabla_{\mathbf{x}} \log p_g(\mathbf{x}).
$$
\end{proposition}

$\nabla_{\mathbf{x}} \log p_*(\mathbf{x})$ represents the score function of the real samples' distribution, and $\nabla_{\mathbf{x}} \log p_g(\mathbf{x})$ represents the score function of the synthetic samples' distribution. 
Thus, training the CFG with the specified loss function boils down to identifying an optimal 
$D(\mathbf{x})$. 
The gradient of this optimal 
$D(\mathbf{x})$ distinguishes between the integral of the difference between $ \nabla_{\mathbf{x}} \log p_*(\mathbf{x})$ and $\nabla_{\mathbf{x}} \log p_g(\mathbf{x})$.

\vspace{1mm}
\noindent\textbf{CFG Generator and Score Function Interrelation.
}
Following our explanation of the relationship between the score function and the gradient of the discriminator, we investigate how the generator's loss function is connected to the score function using the discriminator's gradient.

\begin{proposition}\label{p:2}
$p_*(\mathbf{x})$, $p_g(\mathbf{x})$, and $p_{g_m}(\mathbf{x})$ denote the distribution of real samples, synthesis samples, and the $m$-th step synthesis samples, respectively.
$\mathbf{x}\triangleq G(\mathbf{z})$ where $\mathbf{z} \sim \mathcal{N}(\mathbf{x} \mid \mathbf{0}, \mathbf{I})$, $\delta_m(\mathbf{x}) \triangleq s_m(\mathbf{x})v_m(\mathbf{x})$.
Therefore, the generator's loss function can be written as 
$$\min \sum_i^N \frac{1}{2}\left\|\sum_{i=1}^M \delta_m(\mathbf{x})\left(\nabla_{\mathbf{x}} \log p_*(\mathbf{x}) -  \nabla_{\mathbf{x}} \log p_{g_m}(\mathbf{x})\right)\right\|^2.$$
\end{proposition}

In proposition~\ref{p:2}, the loss function's form shows that training the generator in the CFG attempts to establish an optimal generator that minimizes the score function between the distribution of real and synthesized samples.
Ultimately, this enables the generator to produce realistic samples.

\begin{figure*} [t!]
	\centering
       \subfloat[\label{fig:gradient-detail-church.}]{
		\includegraphics[scale=0.5]{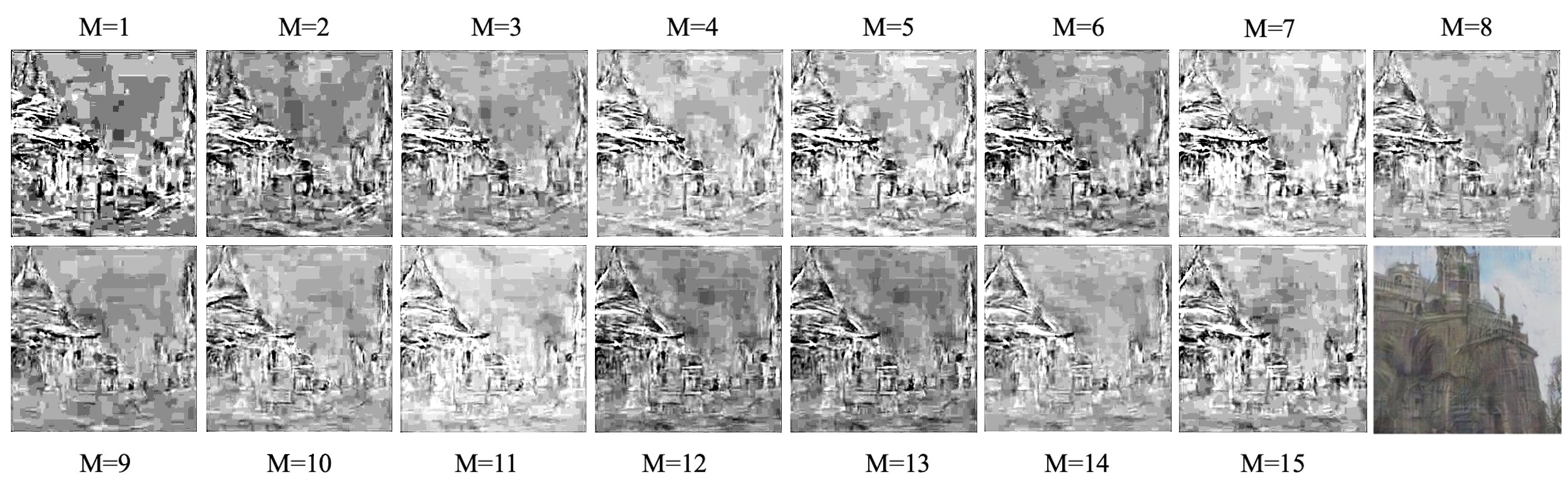}}\\
	\subfloat[\label{fig:gradient-detail-church-anneal}]{
		\includegraphics[scale=0.5]{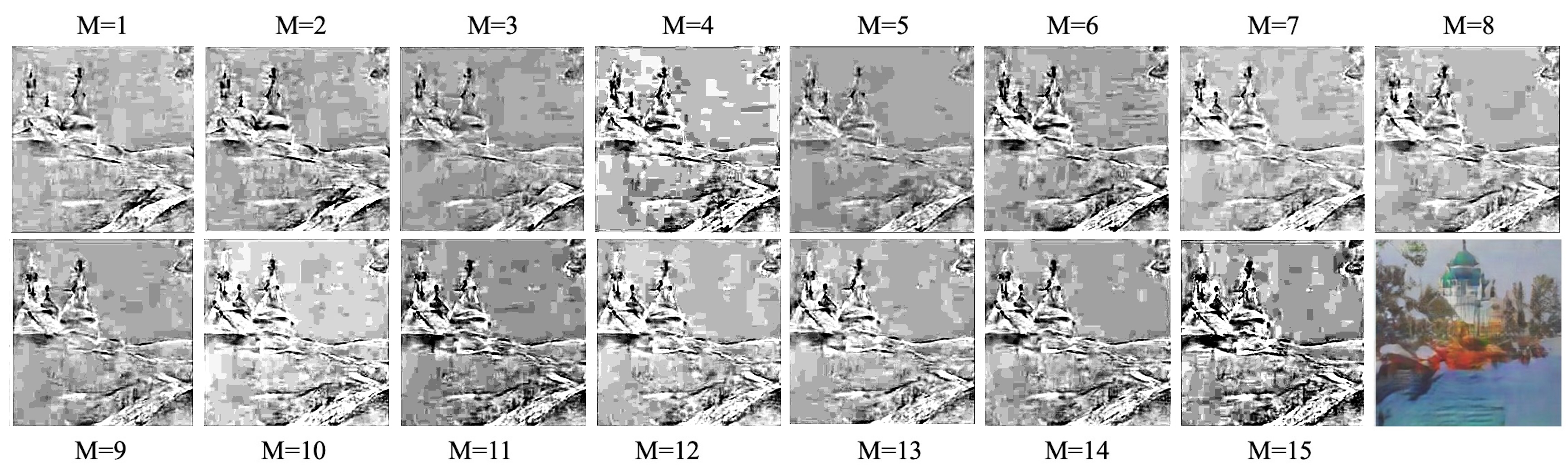}}
	\caption{(a) A failure synthesis sample of CFG and gradient of discriminator in each $M$ Step in 256$\times$256 resolution LSUN Church.
 (b) Synthesized sample of Annealed CFG and gradient of discriminator in each $M$ Step in 256$\times$256 resolution LSUN Church 
The visual effect in the $M=1$ column is more ambiguous than in the $M=15$ column. 
 In the CFG method, each $M$ step shares the same gradient weight. 
 When these gradients accumulate together, the features in $M=1$ will be covered by features in $M=15$. 
 %
The performance of our Annealed CFG has been improved because a geometry annealed weight is set for each M step, with a condition that $\frac{\mathbf{w}_{1}(\mathbf{x})}{\mathbf{w}_{2}(\mathbf{x})}=\cdots=\frac{\mathbf{w}_{m-1}(\mathbf{x})}{\mathbf{w}_{m}(\mathbf{x})} > 1$, with $M=15$, $\mathbf{w}_{1}(\mathbf{x})=20$, and $\mathbf{w}_{15}(\mathbf{x})=1$.
 Each $M$ step's gradient is aligned to maximize the retention of features in each $M$ step.
 The results agree implicitly with the score functions. 
 In the initial steps of the score function, 
 the distribution gradient is ambiguous, and thus we can use a large step size (a larger weight). 
 When the steps come closer to the read data distribution, the distribution gradient is clear, and thus we should be careful and take a smaller step size (a smaller weight).} \label{figure:gradient-detail}
\end{figure*}

\subsection{Annealed CFG}
\label{anneal}

After revealing the nuances of the discriminator gradient in the CFG method, a new question arises: How does this discovery impact CFG methods' training?

We find that if the discriminator's gradient reveals the discrepancy between the scores of real and synthetic sample distributions, it will consequently guide the generator to minimize this gap. 
Taking inspiration from simulated annealing sampling ~\cite{neal2001annealed,kirkpatrick1983optimization,song2019generative}, we theoretically obtain an annealed weight from the CFG discriminator's gradient.
This annealed weight will ideally steer the generator toward aligning more closely with real samples' distribution. 
With this aim in mind, we present our annealed CFG algorithm in Algorithm~\ref{alg:anneal cfg}. 
In this algorithm, we introduce an annealed weight coefficient $\mathbf{w}_{m}(\mathbf{x})$ before the $\nabla_{\mathbf{x}}D(\mathbf{x})$ in Line 5 while keeping all other aspects of the CFG methods the same.
\begin{proposition}
The analytic form of $g_m(\mathbf{x})$ in our annealed CFG model is expressed as follows:
\begin{small}
\begin{gather}\label{eq:anneal g p}
\begin{aligned}
g_m(\mathbf{x}) =\sum_{i=1}^N \delta_{m}^{(i)}(\mathbf{x})r_{m}^{(i)}(\mathbf{x})\nabla_{\mathbf{x}} D_m^{(i)}(\mathbf{x})=\sum_{i=1}^N \mathbf{w}_{m}^{(i)}(\mathbf{x})\nabla_{\mathbf{x}} D_m^{(i)}(\mathbf{x}),
\end{aligned}
\end{gather}    
\end{small}
\end{proposition}
where $\delta_{m}^{(i)}(\mathbf{x})$ refers to the hyper-parameters from CFG, and $r_{m}^{(i)}(\mathbf{x})$ is a coefficient from the score-based model and $\mathbf{w}_{m}^{(i)}(\mathbf{x})=\delta_{m}^{(i)}(\mathbf{x})r_{m}^{(i)}(\mathbf{x})$.

We can update the CFG generator in ~(\ref{eq:generator}) with this annealed formula, leading it to achieve more effective results. 
To show the impact of varying weights on the gradient of the discriminator, we present visual effects in Fig.~\ref{figure:gradient-detail}. 
This figure elucidates the annealed weight's mechanism.

However, certain important questions remain unanswered, which are critical to the success of annealed CFG in such datasets: (1) Is an initial weight of $\mathbf{w}_{1}(\mathbf{x})=1$ suitable for $g_1(\mathbf{x})$ in all datasets? 
If not, how should we adjust the initial weight $\mathbf{w}_{1}(\mathbf{x})$ for different datasets? 
(2) Is the descent progression of annealed weight $\mathbf{w}_{m}(\mathbf{x})$ a good choice?


To answer the above-mentioned questions, we present the following theoretical analysis and empirical observations.

\vspace{1mm}
\noindent\textbf{Initial Annealed Weight of Discriminator.}
\label{anay:init weight}
From ~(\ref{eq:anneal g p}), we see that the annealed weight $\mathbf{w}_m(\mathbf{x})$ for $\nabla_{\mathbf{x}}D(\mathbf{x})$ is  $\delta_{m}^{(i)}(\mathbf{x})r_{m}^{(i)}(\mathbf{x})$.  
When $m=1$, $\delta_{m=1}(\mathbf{x})$ is usually set as a constant corresponding to the training data. 
Referring to principles from score-based models, we know that $r_{m}^{(i)}(\mathbf{x})$ should approximate 0.5, as $\sigma_1$ should be as large as the maximum Euclidean distance among all pairs of training data points. 
Thus, amalgamating these insights, we propose the following technique for selecting $\mathbf{w}_{1}(\mathbf{x})$:
\begin{technique}
Using the median pairwise distance $D$ in the training data as a reference, we set $\mathbf{w}_{1}(\mathbf{x})$ as a smaller value when $D$ is small and opt for a larger value for $\mathbf{w}_{1}(\mathbf{x})$ when $D$ is large.
\end{technique}
We know that $\mathbf{w}_{1}(\mathbf{x}) = \delta_{1}^{(i)}(\mathbf{x})r_{1}^{(i)}(\mathbf{x})$ is made up of two components.
The first component, $r_{1}^{(i)}(\mathbf{x})$, should ideally be maximized to approach 0.5, as indicated by score-based model principles.
The second component is $\delta_{1}(\mathbf{x})=s_{1}(\mathbf{x})v_1(\mathbf{x})$.
The scale function $s_1(\mathbf{x})$ is customarily set to 1. 
Here, we focus on examining $v_1(\mathbf{x})=\tilde{k}(\mathbf{x}) f^{\prime \prime}(\tilde{k}(\mathbf{x}))$. Different function choices of $f$ refer to various distance measures, such as KL divergence and JS divergence. 
For our analysis, we focus on KL divergence with  $f=-\ln x$, allowing us to express 
$v_1(\mathbf{x})$ as follows:
\begin{gather}
\begin{aligned}
v_1(\mathbf{x}) &=\tilde{k}_1(\mathbf{x}) f^{\prime \prime}(\tilde{k}_1(\mathbf{x}))\\
&=\frac{1}{\tilde{k}_1(\mathbf{x})} = \frac{p_*(\mathbf{x})}{p_{g_1}(\mathbf{x})},
\end{aligned}
\end{gather}
where $f^{\prime \prime} = 1/x^2$ and $\tilde{k}_1(\mathbf{x}) \approx \frac{p_{g_1}(\mathbf{x})}{p_*(\mathbf{x})}$.
Assuming that both $p_*(\mathbf{x})$ and $p_{g_1}(\mathbf{x})$ follow what is akin to a normal distribution, we can represent their analytic forms as
$$
p_*(\mathbf{x})=\frac{1}{(2 \pi)^{\frac{d}{2}}|\Sigma_{*}|^{\frac{1}{2}}} e^{-\frac{1}{2}(\mathbf{X}-\mathbf{\mu}_{*})^T \Sigma_{*}^{-1}(\mathbf{X}-\mathbf{\mu}_{*})}
$$ and 
$$p_{g_1}(\mathbf{x})=\frac{1}{(2 \pi)^{\frac{d}{2}}|\Sigma_{g_1}|^{\frac{1}{2}}} e^{-\frac{1}{2}(\mathbf{X}-\mathbf{\mu}_{g_1})^T \Sigma_{g_1}^{-1}(\mathbf{X}-\mathbf{\mu}_{g_1})}.$$
Thus, we can expand the analytic formula for $v_1(\mathbf{x})$ as
\begin{gather}
\begin{aligned}
v_1(\mathbf{x})&=\frac{p_*(\mathbf{x})}{p_{g_1}(\mathbf{x})} \\
&=\frac{1}{|\Sigma_{*}|^{\frac{1}{2}}}e^{\frac{1}{2}(\mathbf{X}-\mathbf{\mu}_{g_1})^T \Sigma_{g_1}^{-1}(\mathbf{X}-\mathbf{\mu}_{g_1})-\frac{1}{2}(\mathbf{X}-\mathbf{\mu}_{*})^T \Sigma_{*}^{-1}(\mathbf{X}-\mathbf{\mu}_{*})}\\
&=\frac{1}{|\Sigma_{*}|^{\frac{1}{2}}}e^{\frac{1}{2}(\mathbf{X})^T(\mathbf{X})-\frac{1}{2}(\mathbf{X}-\mathbf{\mu}_{*})^T \Sigma_{*}^{-1}(\mathbf{X}-\mathbf{\mu}_{*})},\\
\end{aligned}
\end{gather}
where
$p_{*}^{(i)}(\mathbf{x}) \triangleq \mathcal{N}\left(\mathbf{x} \mid \mathbf{x}^{(i)}, \Sigma_{*}\right)$ and 
$p_{g_1}^{(i)}(\mathbf{x}) \triangleq \mathcal{N}\left(\mathbf{x} \mid \mathbf{0}, \mathbf{I}\right)$.
Clearly, 
$$
\frac{1}{2}(\mathbf{X})^T(\mathbf{X})-\frac{1}{2}(\mathbf{X}-\mathbf{\mu}_{*})^T \Sigma_{*}^{-1}(\mathbf{X}-\mathbf{\mu}_{*})<0
$$ and
$$
e^{\frac{1}{2}(\mathbf{X})^T(\mathbf{X})-\frac{1}{2}(\mathbf{X}-\mathbf{\mu}_{*})^T \Sigma_{*}^{-1}(\mathbf{X}-\mathbf{\mu}_{*})} \in [0,1].
$$ 
Given different datasets, the value of  $v(\mathbf{x})$  is primarily influenced by $|\Sigma_{*}|^{-\frac{1}{2}}=\sigma_{*}^{-\frac{N}{2}}$, where $\sigma_{*}\ll 1$, and $N$ represents the dimension of the dataset. 
For example, in CIFAR10, $N$ is $32\times 32$, and in LSUN Tower 64$\times$64, $N$ is $64 \times 64$. 
Thus, the value of $|\Sigma_{*}|^{-\frac{1}{2}}$ or CIFAR10 is much smaller than that for the 64$\times$64 resolution LSUN Tower. 
We can therefore infer that $v(\mathbf{x}) \propto D_2$, defined in Table~\ref{table:pairD}.
Therefore, when the distance between the real distribution and the synthetic sample distribution is large, the value of 
$v(\mathbf{x})$ should be correspondingly large.

\begin{table}\ra{1.3}
\caption{\centering 
We showcase the median pairwise distances between samples across different datasets. 
Here, $D_{1}$
represents the median pairwise distance among the training data, while $D_{2}$
denotes the median pairwise distance between the training and initial synthesized samples.
\label{table:pairD}}
\centering  
\begin{tabular}{@{}m{0.2\textwidth}<{\centering}
m{0.08\textwidth}<{\centering}m{0.08\textwidth}<{\centering}@{}} 
\toprule
\bf DataSets  &\bf $D_{1}$ &\bf $D_{2}$\\
\midrule
CIFAR10 $32\times32$           &38&62  \\
CeleBA $64\times64$       &80&130\\
LSUN Tower $64\times64$     &80& 125  \\
LSUN Church $64\times64$          &75&125  \\
LSUN Bedroom $64\times64$     &76& 124  \\
LSUN Tower $256\times256$       &335& 520\\
LSUN Church $256\times256$       &330& 515\\
\bottomrule
\end{tabular}
\end{table}

\vspace{1mm}
\noindent\textbf{Other Annealed Weight of Discriminator.}
After determining the value of $\mathbf{w}_{1}(\mathbf{x})$, we must define the other elements of $\mathbf{w}_{m>1}(\mathbf{x})$.
In the CFG method, the $\delta_{m}^{(i)}(\mathbf{x})$ should remain constant during training, which implies that $\delta_{1}^{(i)}(\mathbf{x})=\delta_{2}^{(i)}(\mathbf{x})=\ldots=\delta_{M}^{(i)}(\mathbf{x})$. 
Hence, the values of $\mathbf{w}_{m}(\mathbf{x})$ are mainly governed by the variations in 
$r_{m}^{(i)}(\mathbf{x})$.
According to~\cite{song2019generative}, we know that $\sigma_1>\sigma_2>\sigma_3\ldots\sigma_m$ and the $\sigma_i$ and that these $\sigma_i$ values should be chosen to create a geometric progression with a ratio of approximately $\gamma \approx 0.5$.
Given this selection of $\sigma_i$, the values of $r_{m}^{(i)}(\mathbf{x})$ are arranged in descending order such that $r_{1}^{(i}>r_{2}^{(i)}>\ldots>r_{M}^{(i)}$.
Empirically, we also set the  $r_{m}^{(i)}$ as a geometric progression such that $\frac{r_{1}^{(i)}}{r_{2}^{(i)}}=\cdots=\frac{r_{m-1}^{(i)}}{r_{M}^{(i)}}>1$.
%
With the specified choices for  $\delta_{m}^{(i)}(\mathbf{x})$ and  $r_{m}^{(i)}$, we can determine the values for the vector 
$\mathbf{w}_{m}(\mathbf{x})$.
The empirical settings of 
$\mathbf{w}(\mathbf{x})$  are shown in Table~\ref{table:detail settings1}.

By visually presenting the gradients of the discriminator in both CFG and annealed CFG methods in Fig.~\ref{figure:gradient-detail}, we underscore the benefits of selecting $\mathbf{w}(\mathbf{x})$ as an annealed weight.
Our demonstration shows the discriminator's gradient's magnitude differs across different $M$ steps. 
The visual representation shows that at 
$M=1$, the gradient's intensity is relatively weak, containing fewer features. 
Conversely, as $M$ increases, the gradient's intensity is strengthened, including a richer array of features. 
No matter the intensity, all these features play a key role in guiding the generator's training.
If the weight 
$\mathbf{w}(\mathbf{x})$ is uniformly set across all 
$M$ steps, an imbalanced accumulation of 
$\nabla_{\mathbf{x}}D(\mathbf{x})$ can occur, and significant, but weaker, features may be overshadowed by stronger ones.
However, employing an annealed weight 
$\mathbf{w}(\mathbf{x})$, we can assign weaker features greater significance, ensuring that the characteristics of each step's gradient 
$\nabla_{\mathbf{x}}D(\mathbf{x})$ are fully retained in the generator's accumulation process. 
This rigorous consideration of features at each step leads to enhanced results.


\vspace{1mm}
\noindent\textbf{Reverse Variation of Annealed Weight.}
The previous section has suggested that from a theoretical standpoint, the selection of $\mathbf{w}(\mathbf{x})$ should follow an annealing scheme where 
$\mathbf{w}_{1}(\mathbf{x})>\mathbf{w}_{2}(\mathbf{x})>\ldots>\mathbf{w}_{m}(\mathbf{x})$. 
Nevertheless, this raises a related question: Is this theoretical assertion accurate? 
To test this idea, we assess the implications of choosing a reverse annealed weight for $\mathbf{w}(\mathbf{x})$, where 
$\mathbf{w}_{1}(\mathbf{x})<\mathbf{w}_{2}(\mathbf{x})<\ldots<\mathbf{w}_{m}(\mathbf{x})$, and we explore how this alternative approach influences the training results. 
If this reversed annealed approach produces better outcomes than our proposed annealed weighting, our initial theoretical analysis may be flawed. 
To investigate this, we report the results of using this reverse annealed-weight approach for the discriminator's gradient in Table~\ref{table:reverse}.

\vspace{1mm}
\noindent\textbf{Issues with Annealed CFG.}
Although the annealed CFG demonstrates superiority over the standard CFG and achieves more favorable results, applying either the CFG method or annealed CFG to train SOTA GAN models remains a difficult task. 
The CFG method is especially sensitive to the selection of hyper-parameters such as 
$\delta(\mathbf{x})$, $\eta$, and others. 
Also, the training scheme of the SOTA GAN models is often incompatible with the CFG method. 
Using SOTA GAN models within the CFG framework typically leads to sub-optimal outcomes. We showcase these empirical findings in Table~\ref{tab.Counter-intuitively}.

To overcome the CFG model's limitations, we require a new training scheme that incorporates the annealed weight concept from the CFG model and can be adapted to various GAN models. 
To accomplish this, we introduce a novel training scheme in the following section: NATS.


\begin{table}[htb!]\ra{1.3}
\caption{\centering Fr\'{e}chet Distance outcomes for various GAN models using the CFG methods. 
Counter-intuitively, increasing the complexity of the models leads to poorer performance in terms of the FID score.\label{tab.Counter-intuitively}}
\centering
\begin{tabular}{@{}m{0.12\textwidth}<{\centering}
m{0.08\textwidth}<{\centering}m{0.08\textwidth}<{\centering}m{0.08\textwidth}<{\centering}@{}} 
\toprule
{\textbf{DataSets}} & CFG With Resnet& CFG With BigGAN &CFG With SNGAN   \\
\midrule
CIFAR10         &        19.41            &  21.41            &     22.53            \\
CelebA         &       12.52   &       14.45            &       13.58       \\
LSUN Church                    &      11.49                  &         13.92           &   14.53             \\
LSUN Tower                 &   18.83                  &       19.76            &    19.28             \\

\bottomrule
\end{tabular}
\end{table}

\begin{algorithm}[htb]
\caption{Framework for generate $\mathbf{x}$ for Annealed CFG guide the training of the generator. $\mathbf{w}(\mathbf{x})$ is the annealed weight of each corresponding discriminator gradient. $N$ is the total training loop. $M$ is the CFG accumulative steps for the generator. 
$\mathbf{w}(\mathbf{x})$ will be set as a hyper-parameter in empirical training.
\label{alg:anneal cfg}} 
\begin{algorithmic}[1]
\For{$i=1,2, \ldots, N$}
\State
Sample minibatch samples $\left\{\mathbf{z}^{(1)}, \ldots, \mathbf{z}^{(m)}\right\}$ from 
\Statex $\quad\;$ noise prior $p_{\mathbf{z}}$
\State $\mathbf{x} = G_{1}\left(\mathbf{z}\right)$
\For{$m=1,2, \ldots, M$ } 
\State
$g_m(\mathbf{x}) \leftarrow \mathbf{w}_{m}(\mathbf{x})\cdot \nabla_{\mathbf{x}}D_m( \mathbf{x})$ where $\mathbf{w}_{m}(\mathbf{x})$
\Statex $\qquad\ \ $ represents
$\delta_{m}^{(i)}(\mathbf{x})r_{m}^{(i)}(\mathbf{x})$ in Eq.~\ref{eq:anneal g p}
\State
$ G_{m+1}\left(\mathbf{z}\right) =$
$ G_{m}\left( \mathbf{z}\right)+$
$\eta_m g\left(G_{m}\left(\mathbf{z}\right) \right)$, for some
\Statex $\qquad\ \ $  $\eta_m>0$
\EndFor
\State {\bf end for}
\State $\mathbf{x} = G_{M}\left(\mathbf{z}\right)$
\EndFor
\State {\bf end for}
\State \Return $ \mathbf{x}$. 
\end{algorithmic}
\end{algorithm}

\subsection{NATS for GAN}
\label{nested}
This section presents a new training method for GAN, NTS, which utilizes a nested training loop structure for the discriminator and generator. 
This approach entails training the discriminator once and then performing multiple iterations for the generator, instead of the traditional alternating single training iteration for each. 
By adding an annealed weight to the nested training loop, we establish the NATS method. 
This modification ensures that the gradient vector used in updating the generator is aligned with the annealed CFG method, successfully blending nested training with the annealing strategy. 

First, let us we explain the difference between the CTS and NTS methods. 
Second, we discuss why we use annealed weight in the NTS method. 
Third, we offer guidance on selecting a suitable annealed weight for the NATS method.

\vspace{1mm}
\noindent\textbf{Analyzing Discrepancy between CTS and NTS.}
In most GAN models, CTS is used, alternating between training the generator and the discriminator one time each, as shown in Algorithm~\ref{gan}. 
Incorporating the concept of annealed weight into CTS poses several challenges.
First, CTS lacks a sub-loop that can accumulate the 
$\nabla_{\mathbf{x}}D(\mathbf{x})$ effectively to guide the generator.
In CTS, each $\nabla_{\mathbf{x}}D(\mathbf{x})$ is directly correlated with the generator, without the benefit of accumulation.
Second, directly applying an annealed weight to $\nabla_{\mathbf{x}}D(\mathbf{x})$ in CTS creates difficulties in determining the annealed weight's appropriate length. 
This is because the annealing length may need to span the entire duration of the training stages, which is not feasible in most scenarios.

To resolve these challenges, we establish the NTS algorithm, which introduces annealed weight to the GAN training scheme.
This is outlined in Algorithm~\ref{natsgan}. The foundation of NTS is a sub-loop structure in the training regimen, as detailed in Lines 2--10 of Algorithm~\ref{natsgan}. 
This sub-loop mirrors the process described in ~(\ref{eq:generator}), facilitating the accumulation of 
$\nabla_{\mathbf{x}}D(\mathbf{x})$ for the generator. 
This sub-loop is the nested structure, contrasting with the parallel structure found in the CTS.

In the CTS with $K=1$, each loop trains the discriminator and the generator one time each. 
Conversely, in the NTS with $K=1, N_d=4$, the training process is more complex. 
For each  $j = 1,\dots,N_d \text{ where } N_d=4$, the following sequence occurs:
 
1. When $K=1, j=1$, the discriminator is trained once, followed by a single training iteration for the generator.

2. When $K=1, j=2$, the discriminator is trained once, and the generator is trained twice.

3. When $K=1, j=3$, the discriminator goes through one training iteration and three training iterations for the generator.

4. Finally, when $K=1, j=4$, the discriminator is trained once again, after which the generator is trained four times.

This pattern creates an alternating training rhythm between the discriminator and generator, with a gradual increase in the training frequency for the generator.

This training sequence, wherein the discriminator and generator are trained in a nested fashion, makes our NTS approach highly flexible and adaptable. 
Thus, it can be effectively applied to various GAN models, handling structural, loss, or regularization characteristics.



\vspace{1mm}
\noindent\textbf{Employing Annealed Weight for NTS.}
The annealed weight obtained from the discriminator is used during the generator's training phase, as shown in Algorithms.~\ref{natsgan}, Line 10 in NTS. 
Here, the annealed weight vector 
$\mathbf{w}(\mathbf{x})$ is applied in $\mathbf{w}[j]\cdot D\left(G\left(\mathbf{z}^{(i)}\right)\right)$, thus significantly enhancing the GAN training stability. 
After this modification, the NTS can be changed to the NATS method. 

After identifying the differences between the NTS and the NATS, we can now explain the theoretical basis for adding annealed weight to NTS.
\begin{proposition}
The generator within the NATS has a gradient-vector field that is equal to that in the annealed CFG method. 
\end{proposition}
Although differences exist in the loss functions and training schemes of the NATS and the annealed CFG, the gradient-vector field utilized to update the generator in both methodologies has the same analytic formula. 
Hence, from the perspective of dynamic theory, the algorithms for training the generator in NATS and annealed CFG are the same.

\vspace{1mm}
\noindent\textbf{Appropriate Annealed Weight for NATS.}
We can now reveal the selection process for the annealed weight in the NATS. This process involves determining both the initial and subsequent weights in the annealing process. 
As the mechanism of the annealed weight in the annealed CFG has been thoroughly analyzed in Section~\ref{anay:init weight}, the approach for selecting the annealed weight for the discriminator in NATS reflects that in the annealed CFG. 
To demonstrate the effectiveness of this approach, we present the empirical results of SOTA GAN models using different training schemes in Tables~\ref{table:reverse} and~\ref{table:NATSSOTA}.

\begin{algorithm}[htb]
\caption{Nested training scheme of generative adversarial nets (NTS). The number of steps to apply to the discriminator K. The $N_d$ and $N_g$ are a pair of hyper-parameters used in our nested training loop of the discriminator and the generator, respectively.
In particular, the training loop of generator $N_g$ is a sub-inner loop inside the training loop of discriminator $N_d$. After adding an annealed weight $\mathbf{w}$ to each corresponding discriminator gradient, the NTS method becomes the NATS method.
\label{natsgan}} 
\begin{algorithmic}[1]
\For{$i=1,2, \ldots, N$}
\For{$j=1,2, \ldots, N_d$}
\For{$k=1,2, \ldots, K$}
\State
Sample minibatch samples $\left\{\mathbf{z}^{(1)}, \ldots, \mathbf{z}^{(m)}\right\}$  \Statex $\qquad\qquad\,$ from noise prior $p_{\mathbf{z}}$
\State
Sample minibatch samples $\left\{\mathbf{x}^{(1)}, \ldots, \mathbf{x}^{(m)}\right\}$  
\Statex $\qquad\qquad\,$ from data generating distribution  $p_{\text {data }}(\mathbf{x})$  

\State
Update the discriminator by ascending its 
\Statex $\qquad\qquad\,$ stochastic gradient:
\Statex $\qquad\qquad\,$
\begin{footnotesize}
 $
\nabla_{\theta_d} \frac{1}{m} \sum_{i=1}^m\left[\log D\left(\mathbf{x}^{(i)}\right)+\log \left(1-D\left(G\left(\mathbf{z}^{(i)}\right)\right)\right)\right]
$ 
\end{footnotesize}
\EndFor
\State {\bf end for}
\For{$g=1,2, \ldots, N_g$ where $N_g=j$}
\State
Sample minibatch samples $\left\{\mathbf{z}^{(1)}, \ldots,
\mathbf{z}^{(m)}\right\}$ 
\Statex $\qquad\qquad\,$ from  noise prior $p_{\mathbf{z}}$ 
\State
Update the generator by ascending its 
\Statex $\qquad\qquad\,$
stochastic gradient:
\Statex $\qquad\qquad\,$
\begin{footnotesize}
$
\nabla_{\theta_g} \frac{1}{m} \sum_{i=1}^m \log \left(1-\mathbf{w}[j]\cdot D\left(G\left(\mathbf{z}^{(i)}\right)\right)\right)
$
\end{footnotesize}
\EndFor
\State {\bf end for}
\EndFor
\State {\bf end for}
\EndFor
\State {\bf end for}

\end{algorithmic}
\end{algorithm}

\section{Experiments}
\label{sec:experiments}
We present the empirical settings and experimental results related to our annealing mechanism, as well as a comparison with other methods. 
Initially, in the empirical settings part, we describe the training datasets used, specific implementation details, baseline methods for comparison, and the evaluation metrics employed. 
We present the quantitative outcomes achieved by our annealing mechanism and the methods we compared it against based on the predefined evaluation metrics. 
Additionally, we conduct an ablation study to examine the impact of various hyper-parameters within our annealing approach. 
Finally, we showcase the visual quality of synthetic samples generated through our annealing mechanism, highlighting its effectiveness in synthesizing realistic data.

\subsection{Empirical Settings}
\vspace{1mm}
\noindent \textbf{Datasets.} We utilized the CIFAR10 dataset from~\cite{krizhevsky2009learning}, the large-scale scene understanding (LSUN) dataset~\cite{yu2015lsun}, and the CelebA dataset from~\cite{liu2015faceattributes} to conduct a fair comparison with the original CFG and GAN methods. 
Additionally, we selected the 64$\times$64 resolution ImageNet~\cite{deng2009imagenet} as the benchmark for the state-of-the-art GAN model to showcase the effectiveness of our NATS method. 
We specifically opted for the 'church' (LSUN C) and the LSUN Tower (LSUN T) datasets for our experiments due to the limitations of the CFG method in these datasets. 
Furthermore, we presented qualitative results for the LSUN Church, LSUN Tower, and CelebA datasets, all with a resolution of 256$\times$256.

\vspace{1mm}
\noindent \textbf{Implementation Details.}\label{Implementation Details}
To fairly compare our method with the CFG method and other GAN methods, we take the same implementation settings as the compared methods. 
All the experiments were done using a single NVIDIA Tesla v100 or a single NVIDIA GTX 4090. 
The meta-parameter values for the CFG method and other GAN models were fixed to those in Table~\ref{table:meta1}, Table~\ref{table:meta2}, Table~\ref{table:detail settings1} and Table~\ref{table:detail settings2} unless otherwise specified.

\begin{table}\ra{1.3}
\caption{\centering Meta-parameters for CFG methods.\label{table:meta1}}
\centering
\begin{tabular}{@{}p{0.04\textwidth}<{\centering}
p{0.28\textwidth}<{\raggedright}p{0.08\textwidth}<{\centering}@{}} 
\toprule
\bf Name  &\bf Descriptions &\bf Value\\
 \midrule
B        &Batch size of Training data&16/32/64\\
U           &Discriminator update per epoch&1\\
N            &Examples for updating G per epoch&640\\
M            &Number of generating steps in CFG method&15\\
$\mathbf{w}(\mathbf{x})$ &Annealed weight vector for low/high resolution data&[1,0.01]/ [20,1]\\
\bottomrule
\end{tabular}
\end{table}

\begin{table}\ra{1.3}
\caption{\centering Meta-parameters for NATS/NTS training schemes.\label{table:meta2}}
\centering
\begin{tabular}{@{}p{0.04\textwidth}<{\centering}
p{0.28\textwidth}<{\raggedright}p{0.08\textwidth}<{\centering}@{}} 
\toprule
\bf Name  &\bf Descriptions&\bf Value\\
 \midrule
B         &Batch size of Training data&32/64\\
K           &Discriminator update per epoch&1/4\\
$N_d$       &Hyper-parameters for Discriminator&4/5/10/ 15/20 \\
$\mathbf{w}(\mathbf{x})$  &Annealed weight vector&[1,0.01] \\
\bottomrule
\end{tabular}
\vspace{-0.2in}
\end{table}
In Table~\ref{table:meta1} and Table~\ref{table:detail settings1}, we provide the foundational configurations for CFG and Annealed CFG mechanism.
CFG method behaves sensitive to the setting of $\delta(\mathbf{x})$ for generating the image to approximate an appropriate value. 
In our work, we keep the same $\delta(\mathbf{x})$ settings as the CFG method; 
%
For datasets with a sample resolution size less than 
$128\times128$, the meta-parameter annealed weight vector $\mathbf{w}(\mathbf{x})$ decreases from 1 to 0.01 in a geometric progression, denoted by the interval $[1, 0.01]$. 
However, when the sample resolution size escalates to $256\times256$,
$\mathbf{w}(\mathbf{x})$ is accordingly adjusted to $[20, 1]$.
The symbol $\mathbf{w}(\mathbf{x})$ denotes a vector representing the annealed weight applied to the discriminator in the Annealed CFG or NATS. 

The $\delta(\mathbf{x})$ and $\eta$ in Table~\ref{table:detail settings1} stands for the $\delta(\mathbf{x})=s(\mathbf{x})\tilde{k}(\mathbf{x})f_{kl}^{''}(\tilde{k}(\mathbf{x}))$ and learning rate used for optimization methods like Adam in deep learning field respectively. 
To keep the theoretical symbol and our experiment code consistent, we use the same $\delta(\mathbf{x})$, $\eta$, as in the theoretical symbol in this section.

In Table~\ref{table:meta2} and Table~\ref{table:detail settings2}, we provide the foundational configurations for NTS/NATS within various GAN models. 
The significance attributed to symbols $\eta$ and $\mathbf{w}(\mathbf{x})$ aligns with their interpretation in the CFG method. 
The symbol $K$ retains its meaning as elucidated in Algorithm \ref{gan}. 
The notation 
$N_d$, introduced in our NATS framework, represents the 'Nested Number of Discriminators'. 
We have explored $N_d$ values of 5, 10, 15, and 20 across four distinct GAN models, subsequently presenting the IS and FID scores for each. 
In the context of BigGAN and DDGAN, $N_d$ values of 4 and 2 have been chosen, as they yield the most favorable FID and IS metrics in these state-of-the-art models.

\vspace{1mm}
\noindent \textbf{Baselines.}
As a representative of comparison methods, we tested our annealed weight idea from three parts. 

Firstly, we apply our concept of annealed weight to the Composite Functional Gradient GAN (CFG) framework, as described by~\cite{johnson2019framework}. 
The modified version of CFG, which incorporates the annealed weight, is referred to as Annealed CFG in our study.
The pytorch version code of~\cite{johnson2019framework} is from its official hub\footnote{https://github.com/riejohnson/cfg-gan-pt}.

Secondly, we apply our NATS and NTS on various commonly used GAN models to demonstrate their effectiveness. 
It includes
original GAN~\cite{goodfellow2014generative}, 
WGAN (Wasserstein GAN)~\cite{gulrajani2017improved},
LSGAN (Least Squares GAN)~\cite{mao2017least},
HingeGAN~\cite{lim2017geometric}.
%
The PyTorch version code of these four models is from a third-party hub\footnote{https://github.com/LynnHo/DCGAN-LSGAN-WGAN-GP-DRAGAN-Pytorch}.

Lastly, we apply our NATS and NTS on some SOTA GAN models which are known for their ability to generate high-quality, richness diversity images.
BigGAN~\cite{brock2018large}
Introduced by~\cite{brock2018large}, BigGAN represents a state-of-the-art GAN model known for its ability to generate high-quality, high-resolution images. 
The PyTorch version code of~\cite{brock2018large} is from its official hub\footnote{https://github.com/ajbrock/BigGAN-PyTorch}.
DDGAN is a GAN-based model proposed by~\cite{xiao2021tackling}, focusing on employing diffusion processes into GAN for image generation. 
The PyTorch version code of~\cite{xiao2021tackling} is from its official hub\footnote{https://github.com/NVlabs/denoising-diffusion-gan}.

\vspace{1mm}
\noindent \textbf{Evaluation Metrics.} 
Generative adversarial models are known to be a challenge to make reliable likelihood estimates.

\begin{table}\ra{1.3}
\caption{\centering Hyper-parameters of CFG method
\label{table:detail settings1}}
\centering  
\begin{tabular}{@{}p{0.15\textwidth}<{\centering}
p{0.05\textwidth}<{\centering}p{0.05\textwidth}<{\centering}p{0.1\textwidth}<{\centering}@{}}  
\toprule
\bf DataSets  & \bf $\eta$ &\bf $\delta(\mathbf{x})$&\bf $\mathbf{w}(\mathbf{x})$\\
\midrule
CIFAR             &2.5e-4& 1& [1,0.01]\\
LSUN         &2.5e-4& 1& [1,0.01]\\
LSUN $256\times256$     &2.5e-4& 1& [20,1]\\
CelebA         &2.5e-4& 1& [1,0.01]\\
CelebA $256\times256$      &2.5e-4& 1& [20,1]\\
\bottomrule
\end{tabular}
\end{table}

\begin{table}\ra{1.3}
\caption{\centering Hyper-parameters of our NATS.
\label{table:detail settings2}}
\centering  
\begin{tabular}{@{}p{0.17\textwidth}<{\centering}
p{0.1\textwidth}<{\centering}p{0.1\textwidth}<{\centering}@{}}  
\toprule
\multicolumn{1}{c}{\bf Models}  &\multicolumn{1}{c}{\bf $\eta$} &\multicolumn{1}{c}{\bf $N_d$}\\
\midrule
original GAN            &2e-4&5/10/15/20  \\
LSGAN             &2e-4& 5/10/15/20  \\
WGAN             &2e-4& 5/10/15/20  \\
HingeGAN             &2e-4& 5/10/15/20  \\
BigGAN        &2e-4&4\\
DDGAN         &1.25e-4& 5\\
\bottomrule
\end{tabular}
\vspace{-0.2in}
\end{table}

So we instead evaluated the visual quality of generated images by adopting the inception score~\cite{salimans2016improved} and Fr\'{e}chet inception distance~\cite{heusel2017gans}. 
The Inception Score is predicated on the principle that higher-quality generated images will exhibit scores akin to those of authentic images, indicating a parallel in excellence. 
Conversely, the Fréchet Inception Distance quantifies the degree of resemblance between the synthetic images and their real counterparts.
%
Acknowledging the limitations of the Inception Score is crucial, particularly its occasional inability to discern problems such as mode collapse or the omission of specific image modes.
Nonetheless, it generally correlates well with human evaluations of image quality.
%

Additionally, the Fréchet Inception Distance (FID) was utilized as a metric of evaluation.
FID assesses the congruence between the feature representation distributions of authentic data, represented as $p_{\text{data}}$, and those of the synthesized data denoted as $p_g$.
The function $f$ is structured to transform an image into a format compatible with a classifier network's internal representation.
A notable advantage of FID is its propensity to register increased values, indicative of diminished performance, particularly in scenarios of mode collapse where the generator is unable to produce a diverse array of samples. 
However, it is important to recognize that computing FID can be relatively resource-intensive. 
In the ensuing results, these two metrics are frequently referred to as the Inception Score and the Fréchet Inception Distance.

\begin{figure}[h]
\centering 
\includegraphics[width=.5\textwidth]{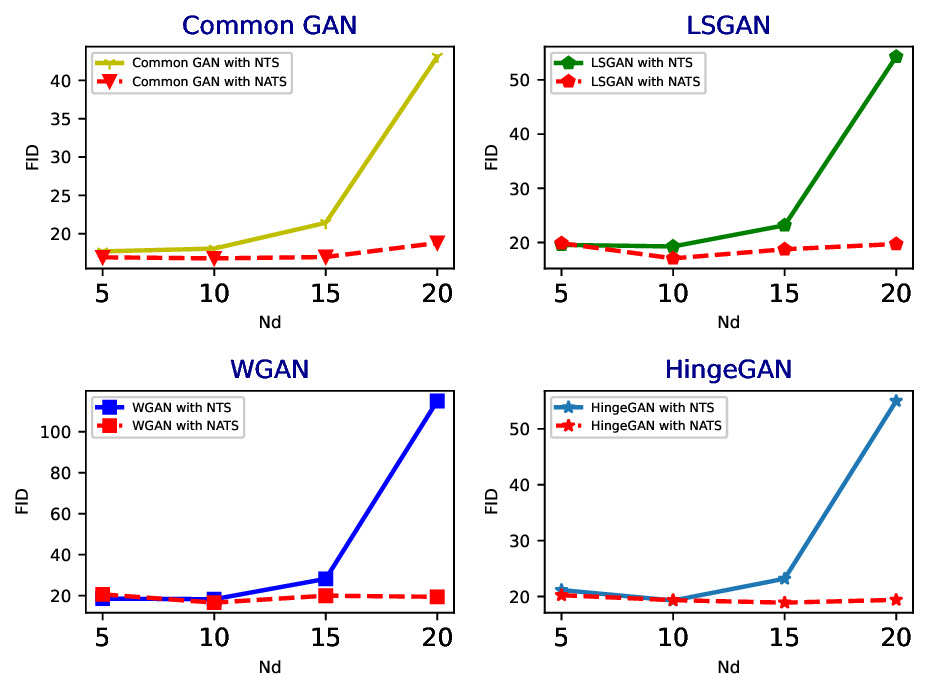}
\caption{Best FID scores for various types of GAN architectures: original GAN, LSGAN, WGAN, and HingeGAN – under different settings for 
$N_d$. The values of $N_d$ considered are 5, 10, 15, and 20. The red line represents the performance of NATS. The other colored lines represent the performance of NTS.
\label{Fig.compare.d}}
\end{figure}

\subsection{Experimental Results}
In this section, we present a comparison of our Annealed CFG and NTS/NATS experimental among the CFG method and other GAN model approaches, respectively. 
We also present the results of our model with the different $\mathbf{w}(\mathbf{x})$, $\delta(\mathbf{x})$ to the CFG and with the different $N_d$ and $\mathbf{w}(\mathbf{x})$ to other GAN model. 
The comparison evaluates that Annealed CFG and NATS archives better result in the approximate meta-parameters.

\begin{table*}[htb!]
\vspace{-0.1in}
\ra{1.3}
\caption{\centering Inception Score Result of CFG and Annealed CFG.\label{table:is} Bigger is better.}
\centering
\begin{tabular}{@{}m{0.1\textwidth}<{\centering}
m{0.1\textwidth}<{\centering}m{0.1\textwidth}<{\centering}m{0.1\textwidth}<{\centering}m{0.1\textwidth}<{\centering}m{0.1\textwidth}<{\centering}m{0.1\textwidth}<{\centering}@{}}
\toprule
\textbf{DataSets}  & WGAN & LSGAN  & Original GAN & HingeGAN  & CFG & Annealed CFG                                                                    \\
\midrule
CIFAR10        & 3.98           & 4.01   &    3.92                                     &    3.89                    & 4.02                                   &                \textbf{4.79}$\dagger$                \\
LSUN B &     2.38                    &       2.31 &       2.33                                   &                  2.42              &  3.02                                  &                \textbf{3.16}$\dagger$                    \\
LSUN T  &  3.67               &      3.52                                        &         3.54                                &         3.62                         &           4.38                             &    \textbf{4.65}$\dagger$                            \\
LSUN C                  &   2.25                &     2.30                                        &     2.23                                    &    2.33                             &               3.17               &      \textbf{3.43}$\dagger$                   
\\
CeleBA                    &   2.23                                        &  2.11                    &  2.18                  &    2.31             &   2.39        &     \textbf{2.43}$\dagger$       \\
\bottomrule                          
\end{tabular}
\end{table*}

\vspace{1mm}
\noindent \textbf{Inception Score Results.} 
We fill the results of the Inception Score (IS) value among different GAN in Table~\ref{table:is}, Table~\ref{table:NATSgeneralization} and  Table~\ref{table:NATSSOTA}.
As the exact codes and models of computing IS scores in~\cite{johnson2018composite} are not available, we use the standard Inception Score function from pytorch~\footnote{https://github.com/sbarratt/inception-score-pytorch}.
Note that the IS scores are affected by many factors, so we recompute all the IS scores in all our experiments with our local computing environments.

\vspace{1mm}
\noindent \textbf{Fr\'{e}chet Distance Results.}
We can see the result of the image quality measured by the Fr\'{e}chet Distance score. 
We use the codes of FID functions\footnote{https://github.com/mseitzer/pytorch-fid}. 
For the sake of impartial comparison, we recalculated the Fr\'{e}chet Distance using 50,000 generative images and all the real images from datasets in all our experiments, utilizing our local computational environments.
%
We demonstrate the results in Table~\ref{table:fid}, Table~\ref{table:NATSgeneralization}, Table~\ref{table:reverse}, Table~\ref{table:NATSSOTA} and Table~\ref{table:ablation}.

\begin{table*}[htb!]
\vspace{-0.1in}
\caption{Fr\'{e}chet Distance results of CFG and Annealed CFG. Smaller is better.\centering}
\label{table:fid} 
\centering
\begin{tabular}{@{}m{0.1\textwidth}<{\centering}
m{0.1\textwidth}<{\centering}m{0.1\textwidth}<{\centering}m{0.1\textwidth}<{\centering}m{0.1\textwidth}<{\centering}m{0.1\textwidth}<{\centering}m{0.1\textwidth}<{\centering}@{}}
\toprule
\textbf{DataSets}  & WGAN & LSGAN  & Original GAN & HingeGAN  & CFG & Annealed CFG\\
\midrule
CIFAR10         & 36.24           & 30.94            &       37.83             &       32.31    &     19.41        &        \textbf{16.34}$\dagger$       \\
LSUN B                      &    18.72    &      19.53                  &        19.68          &  17.62              &     10.79        &        \textbf{5.92}$\dagger$       \\
LSUN T                     &   22.76&   28.14                 &       24.68           &    23.62           &        13.54    &    \textbf{10.79}$\dagger$       \\
LSUN C                    &   32.31                                         &  31.47                     &   30.78                 &   29.87             &    11.49        &     \textbf{7.90}$\dagger$     \\
CeleBA                    &   20.76                                         &  21.14                    &   20.87                 &   20.62              &    12.83        &      \textbf{10.59}$\dagger$     \\
\bottomrule
\end{tabular}
\vspace{-0.1in}
\end{table*}

\begin{figure*} [t!]
	\centering
       \subfloat[\label{fig:biggan_cts}]{
		\includegraphics[scale=0.68]{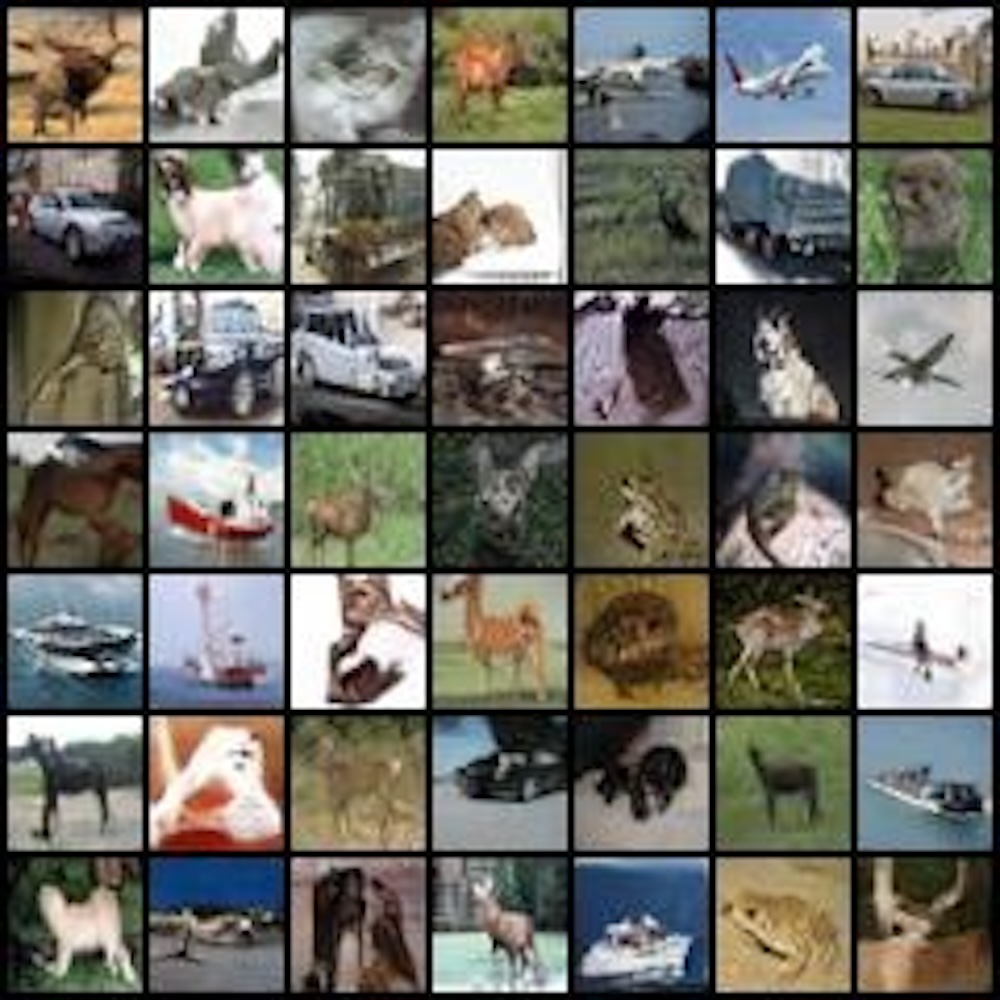}}
	\subfloat[\label{fig:biggan_nts}]{
		\includegraphics[scale=0.68]{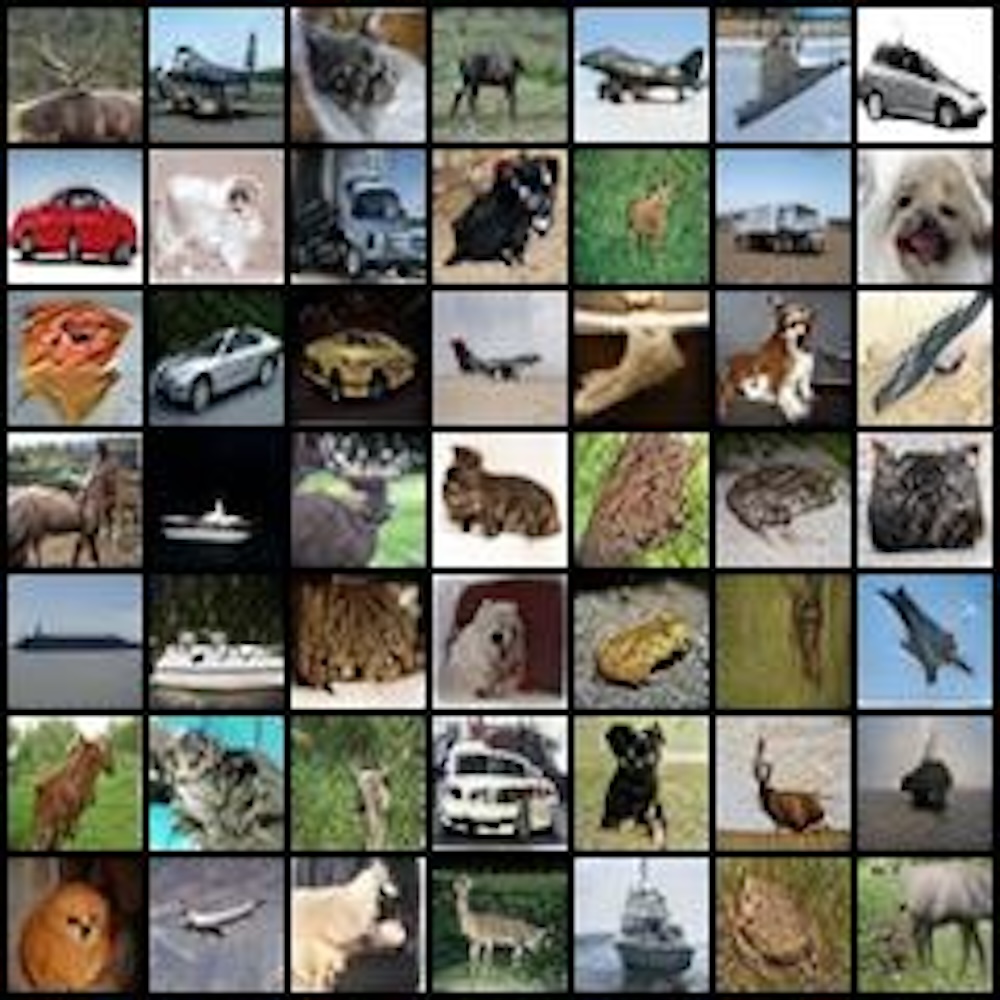}}
       \subfloat[\label{fig:biggan_nats}]{
		\includegraphics[scale=0.68]{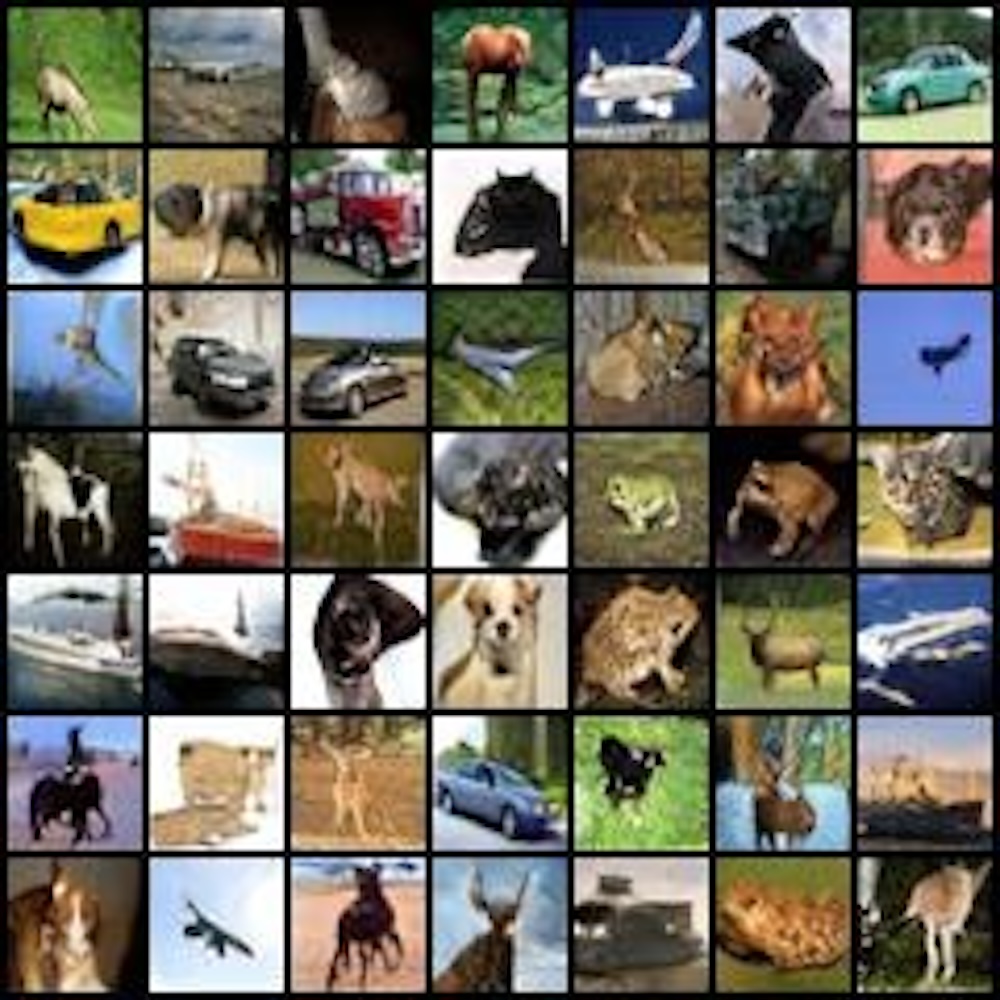} }
	\caption{Performance of BigGAN trained on the CIFAR10 dataset using different training schemes. 
Column (a): This column displays synthetic samples generated by BigGAN trained with the CTS.
Column (b): Here, we present the synthetic samples from BigGAN when trained using the NTS.
Column (c): This column showcases the synthetic samples from BigGAN trained with the NATS.\label{fig.Biggan cifar10} }
\end{figure*}

\begin{figure*} [t!]
	\centering
       \subfloat[\label{fig:dn_cts}]{
		\includegraphics[scale=0.6]{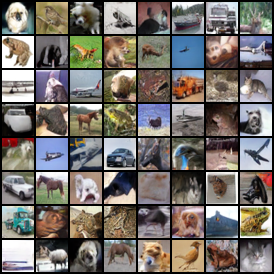}}
	\subfloat[\label{fig:bn_nts}]{
		\includegraphics[scale=0.6]{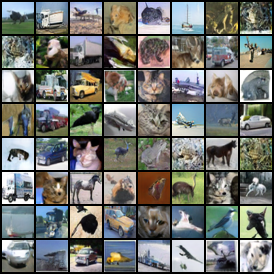}}
       \subfloat[\label{fig:dn_nats}]{
		\includegraphics[scale=0.6]{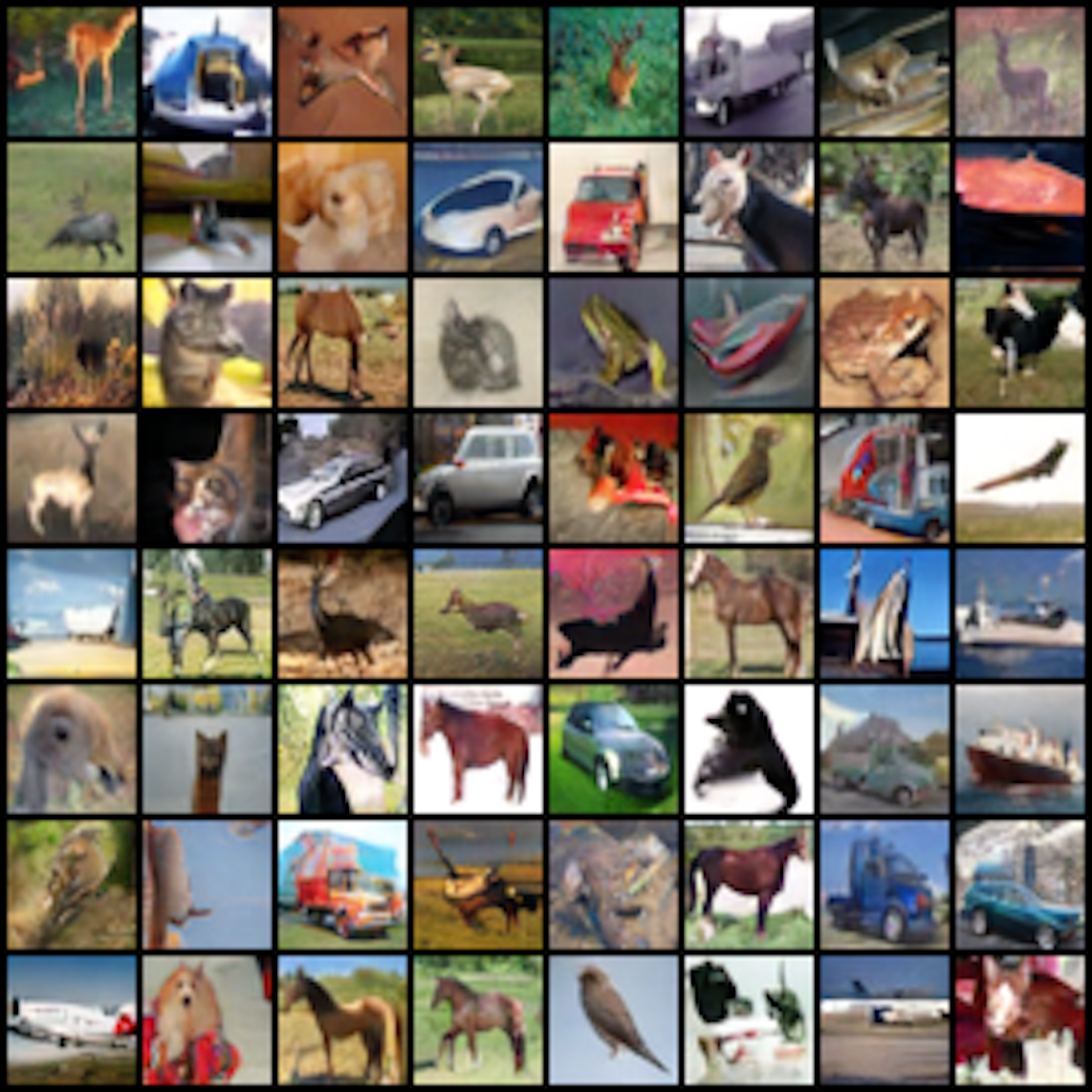} }
	\caption{We present the outcomes of a DDGAN trained on the CIFAR10 dataset using various training schemes.
Column (a): This column displays the synthetic samples produced by the DDGAN when trained with the CTS.
Column (b): Here, the synthetic samples from the DDGAN trained using the NTS are presented.
Column (c): This column showcases the synthetic samples generated by the DDGAN when trained with the NATS.\label{fig.denoising gan cifar10}}
\end{figure*}

\vspace{1mm}
\noindent \textbf{Ablation Study For Hyper-parameters.}
In Fig.~\ref{Fig.compare.d} and Table~\ref{table:ablation}, we investigate the impact of employing various numbers of Nested discriminator $N_d$. 
Note that $N_d$ = 1 and $\mathbf{w}(\mathbf{x})=[1,1]$ corresponds to CTS of GAN model.
With different values of $N_d$ settings, we compare the quality of synthesis samples of the NATS/NTS GAN model using FID and IS scores. 
We demonstrate that our NATS training scheme (each red line) behaviors are more stable and achieve the best FID score than the NTS training scheme in all the $N_d$ settings and models. 
This result is presented in Appendix Section C. 
In Fig.~\ref{Fig.compare.d} and Table~\ref{table:ablation}, we demonstrate that the NATS training scheme achieves the best FID score in all $N_d$ settings. 
However, in WGAN, LSGAN with $N_d=5$ setting, the NATS did not achieve the best FID score. 
That means in the empirical experiments, we should adjust the $N_d$
setting with different GAN loss functions and models. 

In Table~\ref{table:reverse}, we choose a reverse settings of $\mathbf{w}(\mathbf{x})$, from $0.01$ to $1$. 
In the CFG method, the reverse $\mathbf{w}(\mathbf{x})$ setting makes the algorithm diverge to the noise in the CIFAR10 datasets. 
In the BigGAN model, the reverse $\mathbf{w}$ setting also leads to a worse FID result even higher than CTS BigGAN. 
In the original GAN model, the reverse $\mathbf{w}$ setting also led to a worse FID result lower than NTS and CTS but higher than our NATS.

\vspace{1mm}
\noindent \textbf{Visual Synthesis Results.}
We provide the visual synthesis results of our Annealed
CFG in high-resolution datasets. 
We also provide the visual synthesis results of BigGAN and
DDGAN with our NATS on 64$\times$64 resolution ImageNet and 256$\times$256 resolution LSUN Church.
We have included both real images and generated images of a single category using BigGAN with the Common Training
Scheme and our Nest Annealed Training Scheme on CIFAR10 and ImageNet64 datasets.
All these figures are presented in the Appendix Section D.

\begin{table}[htb!]
\ra{1.3}
\caption{\centering We evaluate NATS against CTS, and NTS on several GAN models, including original GAN, LSGAN, WGAN, and HingeGAN using IS and FID scores on the CIFAR10 dataset.
\label{table:NATSgeneralization} 
}
\centering
\begin{tabular}{@{}m{0.1\textwidth}<{\centering}
m{0.03\textwidth}<{\centering}m{0.03\textwidth}<{\centering}m{0.03\textwidth}c<{\centering}m{0.03\textwidth}<{\centering}m{0.03\textwidth}<{\centering}m{0.03\textwidth}<{\centering}@{}}
\toprule
                         &    \multicolumn{3}{c}{FID}       & &\multicolumn{3}{c}{IS}                   \\
                         \cmidrule{2-4}\cmidrule{6-8} 
                &NATS      &  NTS   & CTS  &
            &
           NATS       &  NTS  & CTS     \\
\midrule
original GAN          & \textbf{16.78}$\dagger$          & 17.69  &  37.83 & &\textbf{4.79}$\dagger$                                       &    4.63    &          3.92             \\  
LSGAN        & \textbf{17.08}$\dagger$          &19.28&  30.94&  & \textbf{4.67}$\dagger$                                    &   4.64   &              4.01         \\  
WGAN       &  \textbf{16.59}$\dagger$          & 18.27  & 32.31 & &\textbf{4.79}$\dagger$                                       &    4.66   &  3.98    \\  
 HingeGAN     & \textbf{18.59}$\dagger$           & 19.28  &  36.24 && \textbf{4.75}$\dagger$                                       &    4.57 &     3.89                       \\  
\bottomrule  
\end{tabular}
\end{table}


\begin{table}[htb!]\small
\caption{\centering 
To validate the effectiveness of our geometrically annealed weight approach, we experimented with a reverse annealed weight scheme in the CIFAR10 dataset.
In this reverse weight scheme, the annealed weights are adjusted from a range of [0.01, 1], which is the opposite of the [1, 0.01] range used in our geometric annealing.
When the reverse weight scheme was applied, the FID scores of the synthesized samples were significantly higher, indicating a decrease in image quality and diversity.\label{table:reverse} 
}
\centering
\ra{1.2}
\begin{tabular}{@{}m{0.1\textwidth}<{\raggedleft}
m{0.05\textwidth}<{\raggedleft}m{0.05\textwidth}<{\raggedleft}m{0.05\textwidth}<{\raggedleft}m{0.05\textwidth}<{\raggedleft}@{}}
\toprule
                          &  Reverse      & NATS     & NTS   & CTS                          \\
\midrule
BigGAN         & 16.89          & 6.36 & 7.49  &     8.52                    \\  
CFG        & 200.01         & 16.34   & 19.41      & 19.41                    \\  
original GAN       & 17.59         & 16.92  & 17.69     & 37.93 \\  
\bottomrule
\end{tabular}
\end{table}

\begin{table}[htb!]
\ra{1.3}
\caption{\centering 
To evaluate the generalization capabilities of the NATS, we apply it to prominent GAN models like BigGAN and DDGAN  on the CIFAR10 and ImageNet 64 datasets.
\label{table:NATSSOTA} 
}
\centering
\ra{1.2}
\begin{tabular}{@{}m{0.12\textwidth}<{\raggedleft}
m{0.1\textwidth}<{\raggedleft}m{0.1\textwidth}<{\raggedleft}@{}}
\toprule
             \textbf{CIFAR10}               &    FID       & IS                  \\
\midrule
CTS BigGAN          & 8.52                                            &    8.50                            \\  
NATS BigGAN      &  $\dagger$\textbf{6.36 }       &  $\dagger$\textbf{8.80}                                   \\  
NTS BigGAN       &7.49                       &   8.72  \\
\midrule
CTS DDGAN     & 3.75                                     &     6.59                          \\  
NATS DDGAN          &  $\dagger$\textbf{2.56}                                         &  $\dagger$\textbf{6.81}                      \\ 
NTS DDGAN          & 4.94                                      &    6.63                             \\ 
\midrule 
 \textbf{ImageNet 64}               &    FID        & IS                 \\
\midrule 
CTS BigGAN          &                                 16.5           &    18.12                          \\  
NATS BigGAN      &  $\dagger$\textbf{8.75 }       &  $\dagger$\textbf{25.01}                                   \\  
 \bottomrule
\end{tabular}
\vspace{-0.15in}
\end{table}

\begin{table}[htb!]\small
\caption{\centering 
The FID scores obtained from different 
$N_d$ settings in our NATS applied to various GAN architectures, such as the original GAN, LSGAN, WGAN, and HingeGAN in the CIFAR10 dataset.
It becomes evident that the choice of $N_d$ is crucial for optimal performance. 
Both exceedingly small and excessively large values of $N_d$ can detrimentally impact the results.
  \label{table:ablation} 
}
\centering
\ra{1.2}
\begin{tabular}{@{}m{0.12\textwidth}<{\raggedleft}
m{0.08\textwidth}<{\raggedleft}m{0.08\textwidth}<{\raggedleft}m{0.08\textwidth}<{\raggedleft}@{}}
\toprule
 & $N_d$
& NATS    
&NTS    \\
\midrule
original GAN        &5    &   16.92 &   17.69   \\
original GAN        &10    & $\dagger$\textbf{16.78}     &   18.06     \\
original GAN         &15   &   16.96   &   21.42   \\
original GAN         &20   &   19.36   &   43.09      \\
WGAN          &5   &20.63  &   18.53    \\
WGAN          &10   &$\dagger$\textbf{16.59}  &18.27   \\
WGAN          &15   &19.97  & 32.31 \\
WGAN          &20   &19.43 & 115.01  \\
LSGAN        &5 & 19.88 &19.57\\
LSGAN        &10 & $\dagger$\textbf{17.08}  & 19.28 \\
LSGAN        &15 & 18.76& 23.31\\
LSGAN        &20 & 19.73 & 54.27\\
HingeGAN        &5 & 20.23 &21.15\\
HingeGAN        &10 & 19.34 & 20.14\\
HingeGAN        &15 & $\dagger$\textbf{18.9} & 23.18\\
HingeGAN        &20 & 19.4& 56.72\\
\bottomrule
\end{tabular}
\vspace{-0.2in}
\end{table}

\subsection{Limitation}
In our Annealed CFG method, we established that the discriminator training effectively discerns the discrepancy between the score functions of real and synthetic sample distributions.
However, in the Nested Annealed Training Scheme, our focus was primarily on the gradient vector field of the generator, while the analysis of the discriminator's form was not as extensive. 
This was partly because the analytic solution of the discriminator of Annealed CFG's loss function aligns closely with the theoretical construct of the difference between the score functions of real and synthetic distributions. 
In contrast, other loss functions and their analytic solutions in more general terms might not align as neatly with this theoretical framework.
The lack of a comprehensive theoretical analysis for the discriminator in NATS is indeed a limitation. 
Despite this, empirical results have shown that NATS is effective when applied to SOTA GAN models, demonstrating its practical utility.

Furthermore, our exploration into the optimal values for $M$ and $N_d$ in both Annealed CFG and NATS, particularly for higher resolution datasets, was constrained by the limitations of our hardware computing resources. 
Intuitively, it seems likely that higher-resolution datasets would benefit from larger values of 
$M$ or $N_d$, potentially as high as 
$50 $ or $100$. 
This is because a more extended training sequence allows the discriminator to more precisely distinguish the differences between the score functions of real and synthetic sample distributions. 
However, such an approach would demand significantly more computational resources beyond our available capacity.



\section{Related Work}

\subsection{Score-based model}
The score-based model, a subset of diffusion models, was highlighted by~\cite{song2019generative,song2020improved}, who contrasted it with Markov chain-based diffusion models~\cite{sohl2015deep}~\cite{ho2020denoising}. 
Afterward, the score-based diffusion model~\cite{song2020score} expanded diffusion proceeds to a continuous time framework, offering a unified view of diffusion models and score-based models.
Furthermore, Analytic-dpm~\cite{bao2022analytic} provided the variance boundary's theoretical estimation of diffusion processes. 
Moreover, diffusion-based models have gained widespread adoption across various fields, from computer vision to graphics. 
These applications include multi-modal synthesis~\cite{ramesh2022hierarchical, ramesh2021zero, rombach2022high, saharia2022photorealistic,10288540}, video generation~\cite{ho2022imagen, ruan2023mm,10415463}, and 3D object generation~\cite{zheng2023locally, stan2023ldm3d}, underscoring these models' versatile, significant role in advancing the SOTA in these areas.

\subsection{Theoretical explanation of GAN}
Previous research on GANs' theoretical explanation can be categorized as game theory, optimization or dynamic theory, metrics between distributions, and generalization error analysis.

\vspace{1mm}
\noindent\textbf{Game Theory.}
~\cite{goodfellow2014generative} first conceived of the GAN model as a two-player game, positing that it would lead to a Nash Equilibrium upon the conclusion of the game.
Later game theory research can be categorized into three subsets~\cite{mohebbi2023games}. 
The first subset involves modifying the game mode. 
Here, researchers replace the GAN's zero-sum game with other types of game theory, and the difficulties in this task have been addressed ~\cite{franci2020generative, zhang2018stackelberg, hsieh2019finding}.
The second subset involves modifying the number of players. 
By extending the concept of using a single pair of generator and discriminator to the multi-agent setting, researchers can transform the two-player game into multiple games or multi-agent games~\cite{rasouli2020fedgan,nguyen2017dual,ke2020consistency,li2017triple,fedus2017many,berthelot2017began}.
The third subset entails modifying GAN's learning method. 
These methods~\cite{grnarova2017online,ge2018fictitious,yan2018image} pair other learning approaches such as fictitious play and reinforcement learning with GAN.

\vspace{1mm}
\noindent\textbf{Optimization or Dynamic Theory.}
Besides game theory, much research has concentrated on optimization or dynamic theory to analyze GAN convergence. 
Research such as that by~\cite{che2016mode, chu2020smoothness, nagarajan2017gradient} has evaluated the tendency of GAN to converge to local optimal points and has proposed methods to guide them toward global optimization.
Moreover, studies like~\cite{roth2017stabilizing, mescheder2018training, gulrajani2017improved} have focused on the gradient-vector field of the discriminator and generator, striving to ensure global convergence.

\vspace{1mm}
\noindent\textbf{Metric Between Distribution.}
The third category of theoretical explanations for GAN focuses on the metrics of the discriminator and generator. 
Research including~\cite{arjovsky2017wasserstein, mao2017least, Tran2017, nowozin2016f} has proposed new metrics for the discriminator, utilizing novel loss functions to enhance its capabilities. 
Likewise, studies such as~\cite{johnson2018composite, lambert2022variational, fan2021variational, gao2019deep, nitanda2018gradient} have focused on new metrics for the generator, using gradient flow to minimize these metrics and thus strengthen the generator's performance.

\vspace{1mm}
\noindent\textbf{Generalization Error Analysis.}
The following studies contained theories analyzed from the machine-learning theory perspective ~\cite{liang2021well,huang2022error,hasan2023error}. 
They have focused on evaluating GAN's generalization ability by calculating the generalization error bound of the GAN models. 

Our annealed CFG and NATS models differ from these theoretical frameworks, establishing a novel theoretical connection between GAN and score functions. 
By applying rigorous mathematical derivations from score-based models, we revealed deeper insights into the internal principles of GAN from a theoretical perspective.

\subsection{Generative Adversarial Nets Training Scheme}

Research on GAN training schemes, although not as thorough, can be broadly categorized into three types: one-stage training, consensus optimization, and alternative training.

\vspace{1mm}
\noindent \textbf{One-Stage Training.} Proposed by~\cite{shen2021training}, the one-stage training method establishes a streamlined training scheme that facilitates efficient training of GAN in a single phase. 
This approach aims to simplify the GAN training process, decreasing complexity and potentially increasing training efficiency.

\vspace{1mm}
\noindent \textbf{Consensus Optimization.} Put forth by~\cite{mescheder2017numerics}, the consensus-optimization strategy seeks to replace the common simultaneous gradient ascent approach, serving as a different way to balance the training dynamics between the discriminator and generator.

\vspace{1mm}
\noindent \textbf{Alternative Training.} Initially developed by~\cite{goodfellow2014generative}, the alternative training scheme entails alternating the training of the discriminator and the generator, often with an imbalanced number of updates for each. 
This approach has been fundamental to the development of GAN. 
Further,~\cite{heusel2017gans} suggested that changing the learning rates between the discriminator and generator can enable smoother convergence.

Our NATS can be regarded as an evolution of the alternative training approach. 
What distinguishes NATS is its nested structure, which incrementally increases the generator's training frequency while keeping a constant training frequency for the discriminator.

\section{Conclusion}
This paper has revealed a connection between the GAN model and score-based models, providing a more solid theoretical framework for understanding GAN. 
The training
of the CFG discriminator aims to identify an
optimal $D(\mathbf{x})$. 
The gradient of this optimal $D(\mathbf{x})$ distinguishes between the integral of the differences between the score functions of real and synthesized samples. 
Conversely, training the CFG generator involves finding an optimal $G(\mathbf{x})$ that minimizes this difference. 
Following this line of reasoning, we
have derived an annealed weight preceding the gradient of the CFG
discriminator called
the annealed CFG method. 
To extend the annealed weight concept to a wider range of GAN models, we have introduced the NATS. 
Theoretically, we have proposed that the gradient-vector field of NATS is equal to that of the annealed CFG, as viewed through the lens of dynamic theory.
Empirically, NATS proves its adaptability across various GAN models, exhibiting compatibility and effectiveness no matter their structural, loss, or regularization attributes.
In experimental results, both our annealed CFG and NATS show significant improvements in the quality and diversity of generated samples compared with the CFG method and the CTS. 
Notably, SOTA GAN models trained with NATS exhibit notable enhancements, surpassing the results of original training schemes.



\bibliography{ref}
\bibliographystyle{IEEEtran}


\begin{IEEEbiography}[{\includegraphics
[width=1in,height=1.3in,clip,
keepaspectratio]{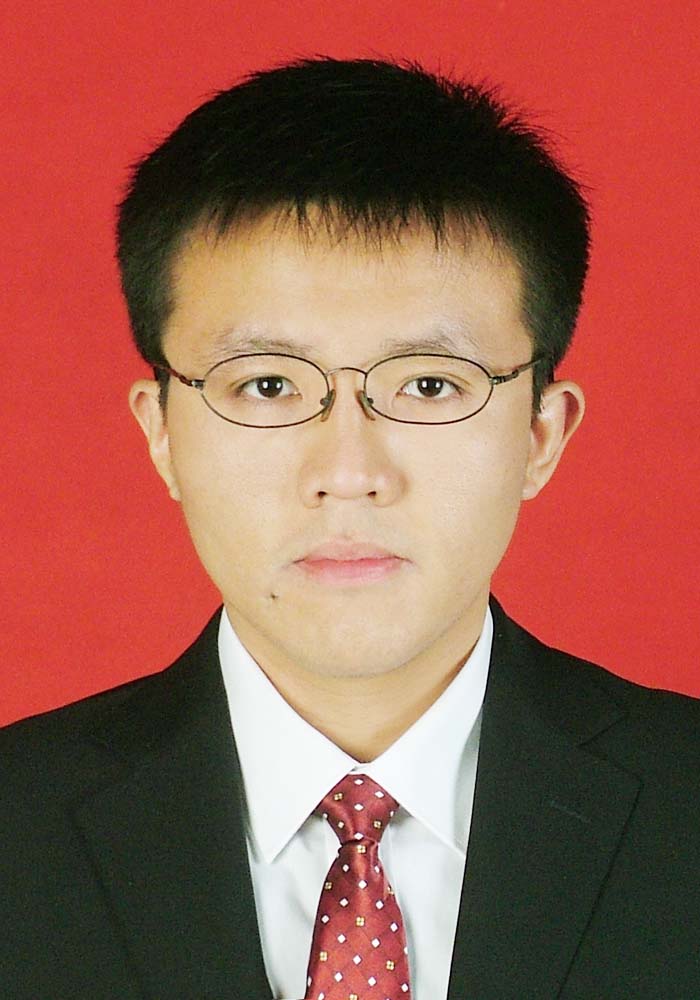}}]
{Chang Wan} Chang Wan received a Bachelor's and Master's degree from Nanchang University, China, in 2009 and 2013, respectively. Now, he is a Ph.D. student at Zhejiang Normal University, China. His research interests include machine learning and computer vision with their applications.
\end{IEEEbiography}
\begin{IEEEbiography}[{\includegraphics
[width=1in,height=1.15in,clip,
keepaspectratio]{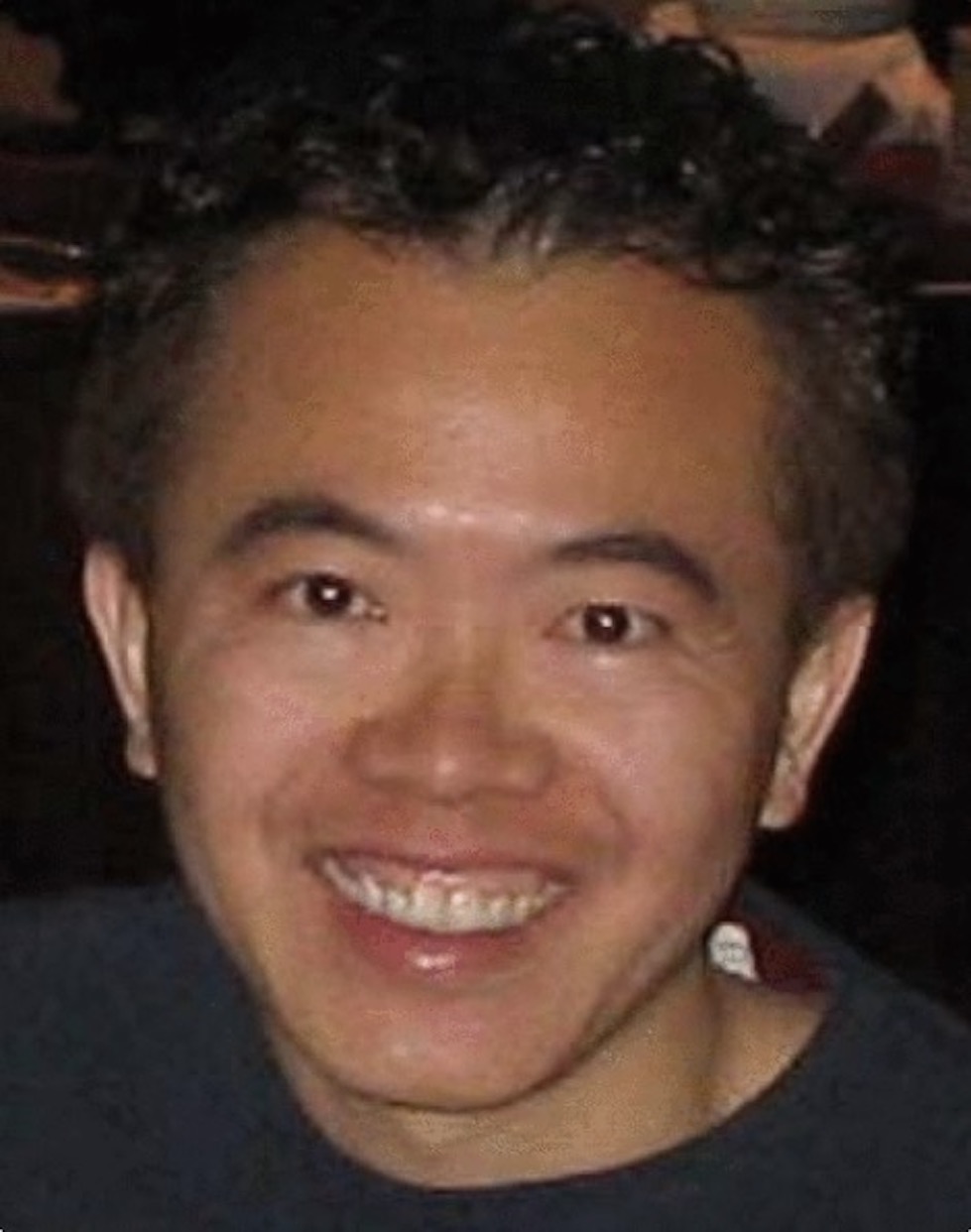}}]
{Ming-Hsuan Yang} Ming-Hsuan Yang (Fellow, IEEE) received the PhD degree from the University of Illinois at Urbana-Champaign. He is a professor of electrical engineering and computer science with the University of California, Merced, and an adjunct professor with Yonsei University. He served as a program co-chair for the 2019 IEEE International Conference on Computer Vision. He served as a co editor-in-chief of Computer Vision and Image Understanding, and an associate editor of the International Journal of Computer Vision. He received the Longuet-Higgins Prize in 2023, NSF CAREER award in 2012 and Google faculty award in 2009. He is a fellow of the ACM.
\end{IEEEbiography}
\begin{IEEEbiography}[{\includegraphics
[width=1in,height=1.25in,clip,
keepaspectratio]{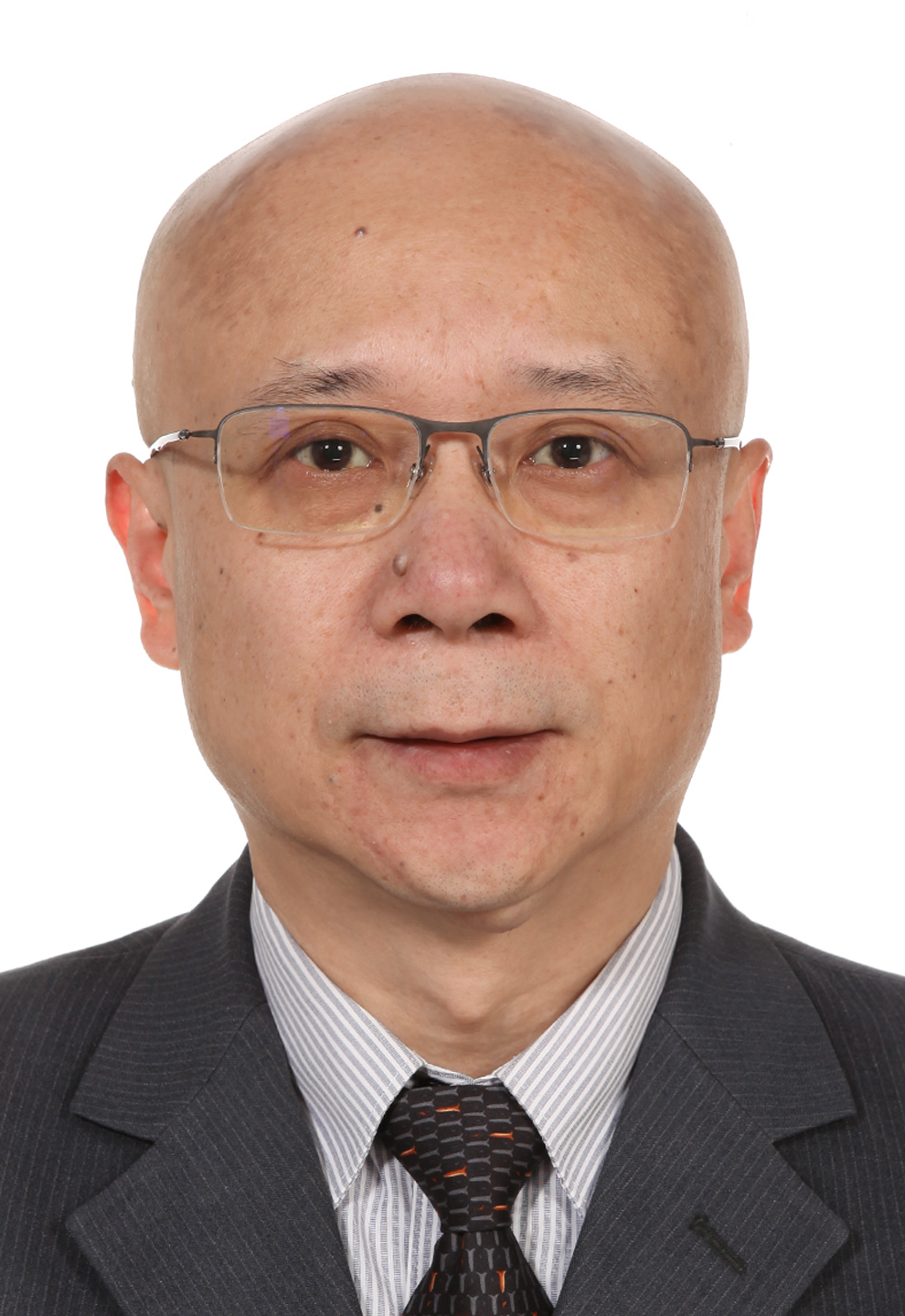}}]
{Minglu Li} Minglu Li (Fellow, IEEE) received the PhD degree in computer software from Shanghai Jiao Tong University, Shanghai, China, in 1996. He is a full professor and the director of the Artificial Intelligence Internet of Things Center, Zhejiang Normal University, Jinhua, China. He is also holding the director of the Network Computing Center, Shanghai Jiao Tong University. He has published more than 400 papers in academic journals and international conferences. His research interests include vehicular networks, Big Data, cloud computing, and wireless sensor networks.
\end{IEEEbiography}
\begin{IEEEbiography}[{\includegraphics
[width=1in,height=1.15in,clip,
keepaspectratio]{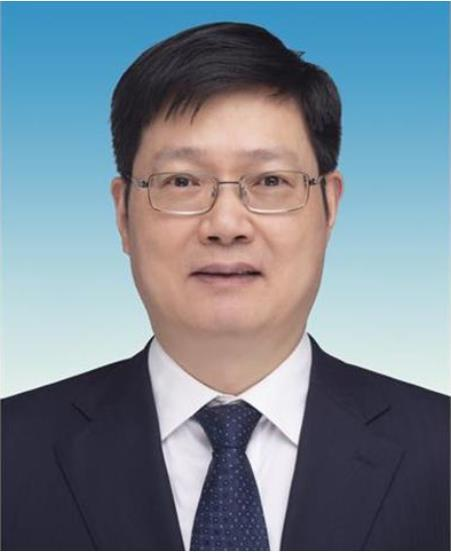}}]
{Yun-Liang Jiang}Yun-Liang Jiang received the Ph.D. degree in computer science and technology from Zhejiang University, Hangzhou, China, in 2006.
He is currently a Professor with the School of Computer Science and Technology, Zhejiang Normal University, Jinhua, China, and also with the School of Information Engineering, Huzhou University, Huzhou, China. His research interests include intelligent information processing and geographic information systems.
\end{IEEEbiography}
\begin{IEEEbiography}[{\includegraphics
[width=1in,height=1.25in,clip,
keepaspectratio]{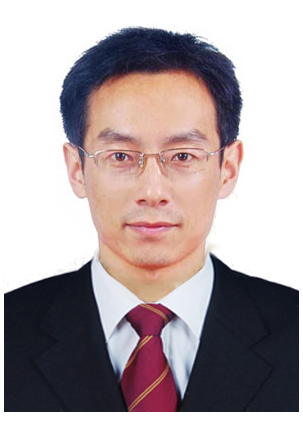}}]
{Zhonglong Zheng} Zhonglong Zheng received the B.Eng. degree from the University of Petroleum, China in 1999, and the Ph.D. degree from Shanghai JiaoTong University, China in 2005. He is currently a full professor and the dean of the School of Computer Science, Zhejiang Normal University, China. His research interests include machine learning, computer vision, and IOT technology with their applications.
\end{IEEEbiography}

\vfill

\end{document}